\documentclass[10pt, a4paper, logo]{googledeepmind} 

\usepackage{textcomp}
\usepackage[normalem]{ulem}
\usepackage{enumitem}

\usepackage{amsmath,amsfonts,bm}
\usepackage{siunitx}

\usepackage{graphicx}
\usepackage{float}
\usepackage{booktabs}
\usepackage{multirow}
\usepackage{tabularx}
\usepackage{array}
\usepackage{threeparttable}
\usepackage{caption}

\usepackage{tikz}
\usetikzlibrary{patterns}
\usepackage{pgfplots}
\pgfplotsset{compat=1.18}

\usepackage{xcolor}
\usepackage{listings}
\usepackage[most]{tcolorbox}
\tcbuselibrary{listings, skins}

\definecolor{jsonbg}{HTML}{F8F9FA}
\definecolor{jsonframe}{HTML}{DFE2E5}
\definecolor{jsonstring}{HTML}{032F62}
\definecolor{jsonnumber}{HTML}{D73A49}
\definecolor{jsonkey}{HTML}{22863A}
\colorlet{punct}{red!60!black}

\lstdefinelanguage{json}{
    basicstyle=\ttfamily\footnotesize\linespread{1.3}\selectfont,
    columns=fullflexible,
    keepspaces=true,
    upquote=true,
    showstringspaces=false,
    breaklines=true,
    breakatwhitespace=true,
    breakautoindent=true,
    breakindent=3em,
    stringstyle=\color{jsonstring},
    morestring=[b]",
    literate=
     *{0}{{{\color{jsonnumber}0}}}{1}
      {1}{{{\color{jsonnumber}1}}}{1}
      {2}{{{\color{jsonnumber}2}}}{1}
      {3}{{{\color{jsonnumber}3}}}{1}
      {4}{{{\color{jsonnumber}4}}}{1}
      {5}{{{\color{jsonnumber}5}}}{1}
      {6}{{{\color{jsonnumber}6}}}{1}
      {7}{{{\color{jsonnumber}7}}}{1}
      {8}{{{\color{jsonnumber}8}}}{1}
      {9}{{{\color{jsonnumber}9}}}{1}
      {\{}{{{\color{black}{\{}}}}{1}
      {\}}{{{\color{black}{\}}}}}{1}
      {[}{{{\color{black}{[}}}}{1}
      {]}{{{\color{black}{]}}}}{1}
      {:}{{{\color{black}{:}}}}{1}
      {,}{{{\color{black}{,}}}}{1},
}

\newtcblisting{jsonbox}[1][]{
    colback=jsonbg,
    colframe=jsonframe,
    boxrule=0.5pt,
    arc=4pt,
    left=3mm, right=3mm, top=3mm, bottom=3mm,
    listing only,
    listing options={
        language=json,
        numbers=left,
        numberstyle=\tiny\color{gray}\ttfamily,
        numbersep=10pt,
        xleftmargin=1.5em,
        aboveskip=0pt,
        belowskip=0pt,
    },
    #1
}

\lstdefinestyle{promptstyle}{
    basicstyle=\ttfamily\footnotesize,
    breaklines=true,
    breakatwhitespace=true,
    frame=single,
    rulecolor=\color{black!30},
    backgroundcolor=\color{black!5},
    xleftmargin=10pt,
    xrightmargin=10pt,
    tabsize=2,
    showstringspaces=false
}

\newcommand{\circlednum}[1]{%
    \tikz[baseline=(char.base)]{
        \node[shape=circle,draw,inner sep=0.5pt,font=\scriptsize,minimum size=1em] (char) {#1};
    }%
}

\usepackage[numbers]{natbib}
\usepackage{url}
\usepackage{hyperref}
\usepackage[capitalize,noabbrev]{cleveref}

\def\eqref#1{equation~\ref{#1}}

% ...

% ...

% ...

\title{QuarkMedBench: A Real-World Scenario Driven Benchmark for Evaluating Large Language Models}

\author{Yao Wu}
\author{Kangping Yin}
\author{Liang Dong}
\author{Zhenxin Ma}
\author{Shuting Xu}
\author{Xuehai Wang}
\author{Yuxuan Jiang}
\author{Tingting Yu}
\author{Yunqing Hong}
\author{Jiayi Liu}
\author{Rianzhe Huang}
\author{Shuxin Zhao}
\author{Haiping Hu}
\author{Wen Shang}
\author{Jian Xu}
\author{Guanjun Jiang}

\affil{Quark Medical Team, Alibaba Group}

\renewcommand{\today}{2026-2-13}

\begin{abstract}

\textbf{Background:} Despite the remarkable performance of Large Language Models (LLMs) on standardized medical examinations such as the USMLE \cite{nori_capabilities_2023, singhal_large_2023, singhal_toward_2025}, high benchmark scores do not necessarily translate to high-quality responses when navigating the multifaceted medical queries of the real world. Current evaluations predominantly rely on structured multiple-choice questions or factual knowledge retrieval, failing to capture the unstructured, ambiguous, and long-tail complexities inherent in genuine user inquiries \cite{liu_large_2024, he_survey_2025}. Consequently, the absence of ecologically valid benchmarks precludes the objective measurement of model efficacy in actual clinical and healthcare deployment.

\vspace{0.5em}
\noindent \textbf{Methods:} To bridge the chasm between evaluation paradigms and practical applications, this study introduces QuarkMedBench, a comprehensive benchmark tailored for real-world medical LLM assessment. First, regarding dataset construction, we employed a multi-axis stratified sampling strategy to compile an evaluation dataset spanning three core dimensions: \textbf{Clinical Care}, \textbf{Wellness Health}, and \textbf{Professional Inquiry}. The benchmark curates a massive-scale dataset comprising 20,821 single-turn queries and 3,853 multi-turn sessions (yielding 6,215 turns), from which 220,617 fine-grained, atomic scoring rubrics were systematically derived (averaging $\sim$9.8 rubrics per query). Second, regarding the evaluation methodology, to overcome the quantitative difficulties of open-ended medical question-answering, we propose a highly efficient and accurate automated scoring rubric generation framework. This method integrates multi-model consensus outputs with external evidence-based retrieval to dynamically generate fine-grained rubrics specific to individual queries. Subsequent blind audits via random sampling by clinical experts verified the absolute medical reliability of these automated standards. Third, in evaluation execution, the system introduces hierarchical weighting and safety constraint mechanisms to structurally quantify the medical accuracy, key-point coverage, and risk-interception capabilities of model responses, thereby effectively mitigating the high costs and inherent subjectivity of traditional human expert evaluations.

\vspace{0.5em}
\noindent \textbf{Conclusions:} This study releases the QuarkMedBench dataset alongside its corollary automated evaluation framework. Experimental results demonstrate that the benchmark not only precisely aligns with genuine online user intents but also exhibits high reliability; the generated rubrics achieved a 91.8\% concordance rate with human expert adjudications. QuarkMedBench establishes a rigorous, reproducible evaluative yardstick for measuring the actual performance of medical LLMs when handling complex health issues. Furthermore, its underlying generation framework inherently supports the timely, dynamic updating of medical knowledge, circumventing the obsolescence typical of traditional benchmarks, and holds substantial generalization potential for extension into other rigorous vertical domains.
\end{abstract}

\begin{document}
\maketitle

\begin{figure}[H]
    \centering
    \definecolor{QuarkBlue}{HTML}{1F4BFF}
    \definecolor{SoftStone}{HTML}{E2E4E9}
    
    \begin{tikzpicture}
    \begin{axis}[
        width=\textwidth,           
        height=0.48\textwidth,      
        ybar=1.5pt,                 
        bar width=9.5pt,            
        enlarge x limits=0.06,      
        ymin=40, ymax=98,           
        ytick={40, 50, 60, 70, 80, 90}, 
        ylabel={\textbf{Score (\%)}},
        ylabel style={font=\small\bfseries, yshift=-5pt},
        symbolic x coords={GPT-5, Gemini 2.5 Pro, Qwen3-30B-Think, DeepSeek-V3.2, GPT-5.1, Kimi-K2, DeepSeek-R1, Gemini 3.0 Pro, Qwen3-235B, GPT-5.2, Qwen3-32B, Doubao-1.6, Qwen3-30B, Qwen3-8B},
        xtick=data,
        xticklabel style={rotate=35, anchor=north east, font=\scriptsize\bfseries, yshift=-2pt},
        nodes near coords={\pgfmathprintnumber\pgfplotspointmeta},
        every node near coord/.append style={
            font=\fontsize{5.5}{6}\selectfont\bfseries,
            color=black,
            rotate=90,
            anchor=west,
            yshift=2pt,
            xshift=0pt,
            /pgf/number format/fixed,
            /pgf/number format/fixed zerofill, 
            /pgf/number format/precision=1
        },
        legend style={
            at={(0.5, 1.08)},          
            anchor=south,
            legend columns=-1,         
            draw=gray!30,              
            fill=white,
            font=\scriptsize\bfseries,
            /tikz/every even column/.append style={column sep=0.5cm} 
        },
        axis line style={draw=gray!40, line width=0.5pt}, 
        ymajorgrids=true,                                 
        grid style={dashed, gray!30},                     
        tick style={draw=none}
    ]

    \addplot[
        draw=gray!80,    
        line width=0.5pt,
        preaction={fill=SoftStone},
        pattern=north east lines,  
        pattern color=QuarkBlue!70 
    ] coordinates {
        (GPT-5, 79.8) 
        (Gemini 2.5 Pro, 74.6) 
        (Qwen3-30B-Think, 74.5) 
        (DeepSeek-V3.2, 71.4) 
        (GPT-5.1, 70.9) 
        (Kimi-K2, 70.8) 
        (DeepSeek-R1, 70.0) 
        (Gemini 3.0 Pro, 69.1) 
        (Qwen3-235B, 67.6) 
        (GPT-5.2, 66.2) 
        (Qwen3-32B, 64.9) 
        (Doubao-1.6, 61.9) 
        (Qwen3-30B, 58.9) 
        (Qwen3-8B, 57.2)
    };
    \addlegendentry{Unconstrained length}

    \addplot[
        draw=QuarkBlue,           
        line width=0.5pt,
        fill=QuarkBlue            
    ] coordinates {
        (GPT-5, 82.2) 
        (Gemini 2.5 Pro, 62.5) 
        (Qwen3-30B-Think, 66.2) 
        (DeepSeek-V3.2, 64.2) 
        (GPT-5.1, 69.0) 
        (Kimi-K2, 71.6) 
        (DeepSeek-R1, 63.9) 
        (Gemini 3.0 Pro, 65.7) 
        (Qwen3-235B, 67.7) 
        (GPT-5.2, 72.5) 
        (Qwen3-32B, 63.6) 
        (Doubao-1.6, 56.1) 
        (Qwen3-30B, 60.1) 
        (Qwen3-8B, 53.6)
    };
    \addlegendentry{Length constraint ($\le$1000 words)}

    \end{axis}
    \end{tikzpicture}
    
    \caption{\textbf{Performance of mainstream LLMs on QuarkMedBench.} This grouped bar chart compares the scores of 14 models without length constraints (hatched bars) versus those with a strict length constraint of $\le$1000 words (solid bars).}
    \label{fig:main_performance}
\end{figure}
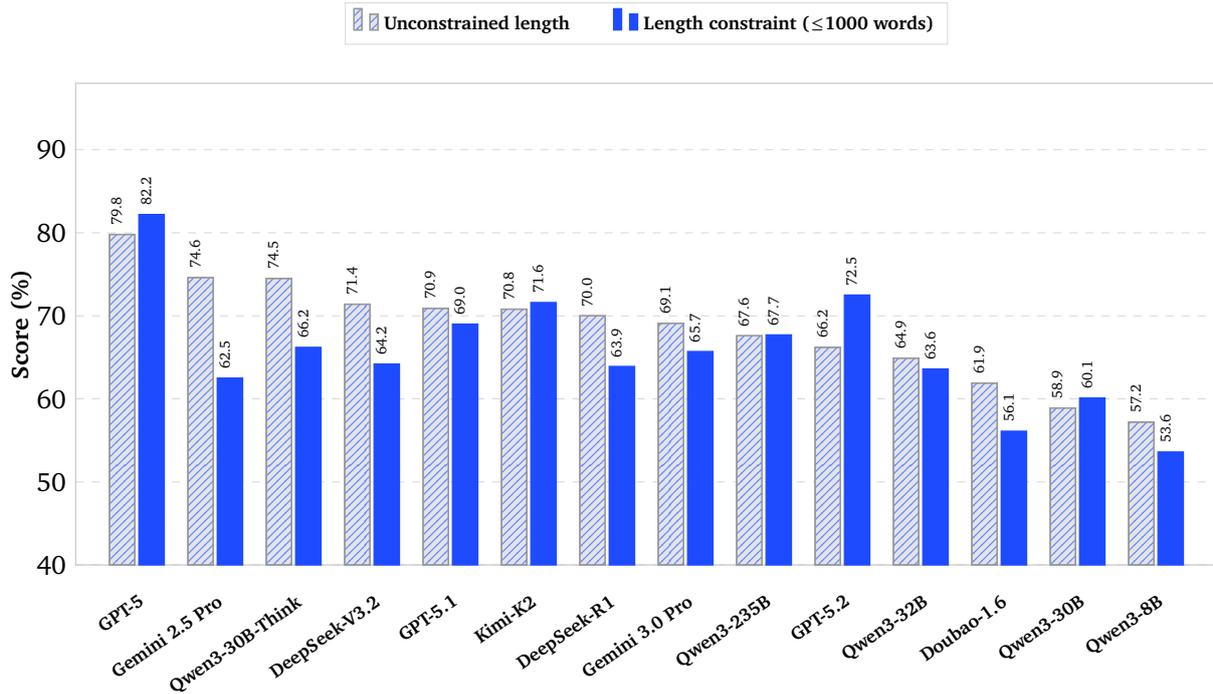

\section{Introduction}
\label{sec:intro}

In recent years, large language models (LLMs) have achieved breakthrough advancements in natural language processing, eliciting profound interest in their application potential for healthcare question-answering (QA) and clinical decision support scenarios. To objectively evaluate the medical knowledge proficiency of LLMs, the academic community has established a series of standardized benchmark tests. For instance, MedQA \cite{jin_what_2021} nd MedMCQA \cite{pal_medmcqa_2022} gauge models' medical knowledge reservoirs by simulating professional examinations such as the United States Medical Licensing Examination (USMLE); PubMedQA\cite{jin_pubmedqa_2019} focuses on evaluating reading comprehension of biomedical literature; whereas MultiMedQA \cite{singhal_large_2023} conducts comprehensive assessments across diverse medical domains. In these standardized tests, frontier LLMs have demonstrated performances that rival or even surpass those of human examinees \cite{nori_capabilities_2023}. For example, both DeepSeek-R1 and ChatGPT-4o achieved an accuracy exceeding 90\% on the Chinese National Medical Licensing Examination (CNMLE) \cite{wu_performance_2025}; the Quark Healthcare LLM similarly passed the chief physician written examinations across 12 core disciplines in China \cite{li_quarkmed_2025}. Notwithstanding their maturation in memorizing and structurally retrieving static medical knowledge, the clinical validity and safety of these models in navigating complex, unstructured real-world health inquiries warrant further empirical validation. Consequently, the evaluation paradigm for medical LLMs is undergoing a systemic paradigm shift---transitioning from rudimentary ``static medical knowledge assessment'' toward ``real-world clinical adaptability evaluation'' \cite{thirunavukarasu_large_2023}.


Aligning with this research trajectory, the industry has recently explored evaluation methodologies predicated on open-ended generation. For instance, OpenAI's HealthBench \cite{arora_healthbench_2025} introduced an evaluation framework based on fine-grained Criterions, employing a foundational criterion set and a difficult criterion set to assess models' foundational medical capabilities and higher-order reasoning proficiencies, respectively. Furthermore, related studies by ScaleAI \cite{gunjal_rubrics_2025} and Ant Group \cite{huang_reinforcement_2025} have corroborated the efficacy of fine-grained rubrics in model value alignment. However, a rigorous analysis of existing evaluation frameworks reveals three primary limitations that impede the deployment of medical LLMs in real-world clinical practice: 
\begin{itemize}
\item \textbf{First, a pronounced distribution shift exists between benchmark data and real-world application scenarios.} Prevalent benchmarks predominantly source data from standard medical textbooks or structured electronic health records from a physician's perspective \cite{thirunavukarasu_large_2023}. Taking HealthBench as an example, its input queries typically comprise high-information-density professional text (averaging 45.4 characters in length), and the scenarios are highly concentrated on intra-consultation treatment recommendations (42.6\%) and medical literature queries (22.58\%) by professional physicians.Conversely, within the authentic Internet healthcare ecosystem, patient-driven health inquiries frequently exhibit highly colloquial, ambiguously intentioned, and long-tail distributions \cite{he_survey_2025}. Our statistical analysis demonstrates that real-world user queries average a mere 11.8 characters, with core intents concentrated on symptom self-checking (13.4\%) and foundational health education (30.7\%). This distribution shift---from structured clinical diagnosis to unstructured patient consultation---frequently renders models that excel on existing benchmarks incapable of providing precise and aligned responses to real patients' health inquiries.
\vspace{0.5em}
\item \textbf{Second, existing evaluation dimensions exhibit inherent limitations.} High scores on objective multiple-choice questions do not linearly translate into the capability to resolve complex, real-world medical problems \cite{singhal_large_2023}. In practical applications, medical QA necessitates not only factual accuracy but also strict adherence to safety baselines, logical consistency, and empathetic engagement with patients. Current evaluations predominantly prioritize the recall accuracy of discrete knowledge points, largely neglecting the model's comprehensive capacity to provide risk warnings, alleviate user anxiety, and maintain explanatory coherence during open-ended generation. 
\vspace{0.5em}
\item \textbf{Finally, current evaluation paradigms suffer from scalability bottlenecks.} To enhance the reliability of open-ended QA evaluation, existing frameworks (e.g., HealthBench) heavily rely on a paradigm where medical experts manually draft scoring standards \cite{arora_healthbench_2025}. This paradigm is not only cost-prohibitive and susceptible to subjective bias but also entails protracted iteration cycles, inevitably precipitating benchmark knowledge decay. Consequently, it fails to adapt to the rapid evolution of medical knowledge and LLM capabilities. The field urgently requires a standardized evaluation framework capable of anchoring expert-level assessment quality while enabling low-cost, scalable updates through automated pipelines.
\end{itemize}

To systematically surmount these bottlenecks, this study proposes \textbf{QuarkMedBench}, a fine-grained evaluation benchmark tailored for real-world medical scenarios. This benchmark provides not only an evaluation dataset with high ecological validity but also an advanced, highly efficient, and accurate methodology for generating scoring rubrics. The primary contributions of this research are summarized in the following three aspects: 

\begin{itemize}[itemsep=0.5em]
\item \textbf{First, construction of a real-world scenario-driven evaluation dataset.} Addressing the disconnect between existing benchmarks and practical applications, this study leverages authentic Internet healthcare logs and employs a stratified sampling strategy to construct a dataset encompassing three core dimensions: rigorous medical consultation, general health inquiries, and professional queries. This design authentically maps the long-tail demand distribution across daily health education, disease diagnosis and treatment, and professional medical queries, effectively calibrating the distribution bias of evaluation data. 
\item \textbf{Second, proposal of an automated, multi-model consensus-based rubric generation framework.} To overcome the bottlenecks of exorbitant manual evaluation costs and delayed updates, this study engineers an algorithmic pipeline for the automated generation of query-sensitive scoring rubrics. This approach aggregates the generative outputs of multiple frontier LLMs to distill core medical knowledge points, automatically constructing fine-grained scoring criteria (Rubrics) tailored to specific queries, and incorporates a random sampling audit by clinical experts. This automated pipeline ensures that the benchmark can undergo high-timeliness dynamic updates in alignment with the latest medical knowledge, effectively circumventing benchmark obsolescence; simultaneously, this rubric generation framework possesses high generalizability, allowing seamless transfer to other high-fault-intolerance vertical domains such as law and finance.
\item \textbf{Third, introduction of a fine-grained quantification mechanism with multi-dimensional constraints.} To resolve scoring weight imbalances and the Length Bias inherent in model generation, this study structures the evaluation dimensions into Essential, Important, Highlight, and Pitfall criteria, assigning differential weights based on clinical criticality. Concurrently, we introduce a truncation mechanism (imposing mandatory penalties for outputs triggering safety redlines or core factual errors) and a saturation mechanism (capping the score for generating homologous redundant information). This mechanism effectively suppresses the propensity of models to achieve high scores via text inflation, ensuring that medical accuracy and clinical safety serve as the definitive cornerstones of the evaluation.
\end{itemize}

\section{Methodology}
\label{sec:methodology}

To address the limitations of traditional human evaluation—namely high cost, long cycles, and poor reproducibility—this study proposes a benchmark construction methodology based on Large Language Models. Although the LLM-as-a-Judge paradigm is widely adopted in general domains, directly relying on a single model to generate scoring rubrics in the zero-tolerance medical field poses significant risks \cite{croxford_automating_2025, zhang_sirens_2025}. On one hand, inherent model hallucinations and style variances can lead to factual errors and inconsistent scoring criteria; on the other hand, real-world health consultations frequently involve long-tail demands or specific localized contexts. Relying solely on the static, parameterized knowledge inside an LLM is insufficient to cover such dynamic knowledge points, resulting in scoring justifications that lack factual support.

Therefore, this study designs a standardized pipeline spanning from reference answer construction to scoring rubric generation, aiming to establish an evaluation benchmark that is simultaneously \textbf{accurate, comprehensive, and objective}. First, the pipeline establishes a reliable Ground Truth (GT). For complex queries involving model knowledge blind spots or ambiguities, external knowledge retrieval technologies (DeepResearch) are introduced to acquire authoritative evidence, thereby correcting cognitive biases and ensuring answer accuracy. Building upon this, the system utilizes LLMs to decouple the verified GT into fine-grained scoring rubrics (Rubrics). We enforce strict control across three dimensions: \textit{Accuracy} (e.g., unambiguous definitions), \textit{Comprehensiveness} (e.g., multi-dimensional capability tags), and \textit{Scoring Logic} (e.g., differential weighting and truncation mechanisms) to establish the reliability and validity of the benchmark. Furthermore, a Human-in-the-loop (HITL) validation mechanism is introduced, refining generation strategies through clinical specialists' audits and feedback, which effectively mitigates the mismatch between model-generated content and genuine clinical needs. This step-by-step construction strategy circumvents the hallucination risks associated with direct generation, balancing automated efficiency with medical factual rigor. Additionally, this study integrates recent advancements in using Rubrics as Rewards for LLM reinforcement learning (RL) training, drawing inspiration from ScaleAI's rubric-guided RL and Ant Group's technical report on Rubric Anchors \cite{gunjal_rubrics_2025, huang_reinforcement_2025}.

\subsection{Evaluation Rubric Generation}
\label{sec:rubric_generation}

\subsubsection{Automated Rubric Generation Pipeline}

\begin{figure}
    \centering
    \includegraphics[width=1.0\linewidth]{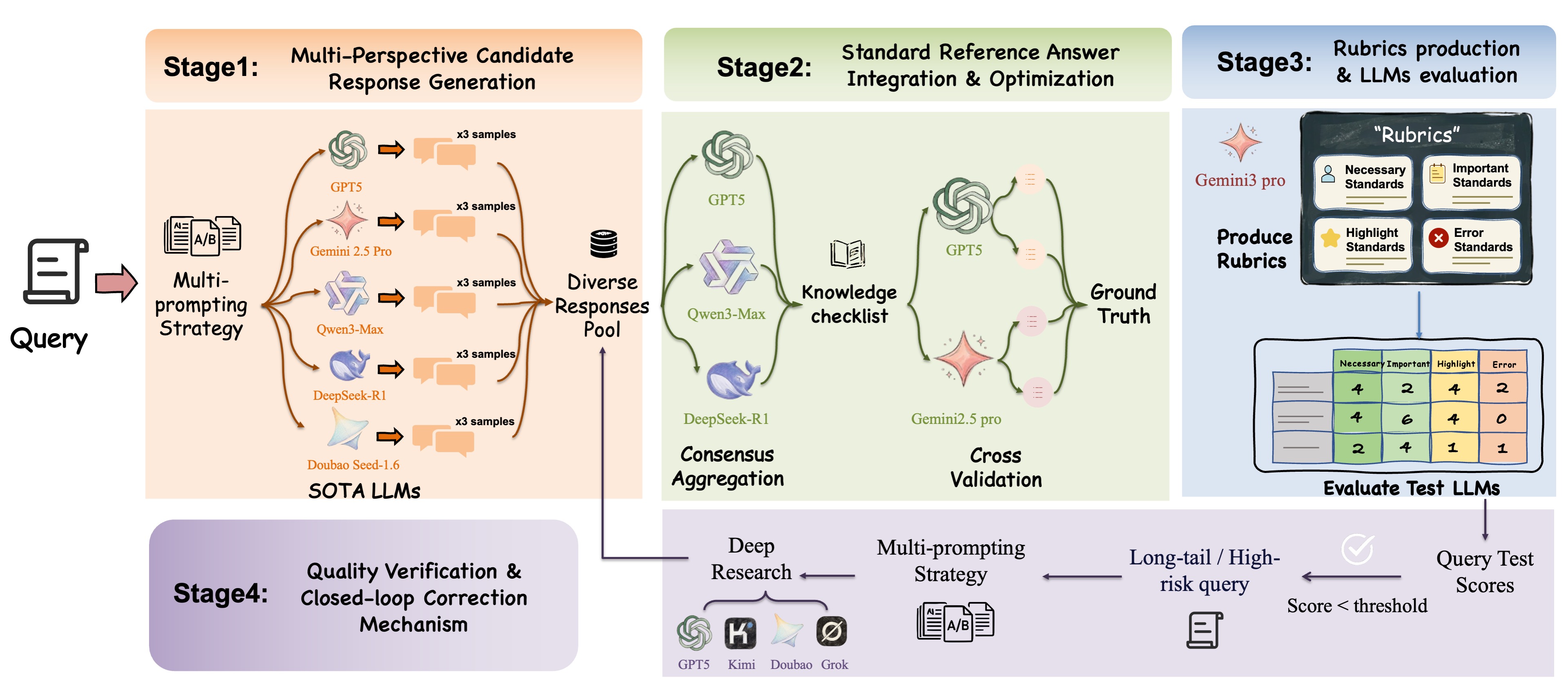}
    \caption{Automated Scoring Rubric (Auto-Rubrics) Generation Pipeline}
    \label{fig:placeholder}
\end{figure}

To address the absence of gold standards and the pervasive subjective bias of human evaluation in open-ended medical question-answering, this study designs and implements a four-stage automated scoring rubric (Auto-Rubrics) generation pipeline. By adopting a systematized paradigm of (1) multi-model generation, (2) knowledge extraction, (3) rubric deconstruction, and (4) closed-loop calibration, this pipeline integrates heterogeneous multi-source information with rigorous quality control mechanisms. The overarching objective is to deterministically transform unstructured healthcare consultation text into an objective, fine-grained, and interpretable quantitative evaluation framework (the architectural flow is illustrated in Figure \ref{fig:placeholder}).


\paragraph{Stage 1: Candidate Response Generation}
To maximize the acquisition of a high-information-density candidate text corpus, subsequent to the retrieval, cleaning, and desensitization of authentic online user queries, this stage employs a multi-model, multi-prompting strategy to construct a diverse candidate response pool. We select five frontier large language models with distinct architectural characteristics (Gemini 2.5 Pro, GPT-5, Qwen3-Max, DeepSeek-R1, and Doubao Seed 1.6) as generation foundations. To capture the maximum diversity of the solution space, each model executes three independent inferences utilizing high-temperature sampling. At the instruction layer, the system designs targeted prompts from three distinct perspectives: foundational cognition (Basic), professional in-depth analysis (Pro), and heuristic innovation (Aha). This directs the models to output multidimensional informational content, ensuring the response pool encompasses the full spectrum of medical information---from general health education to cutting-edge clinical analysis.
\paragraph{Stage 2: Standard Reference Answer Integration}
This stage aims to distill high-confidence Ground Truth (GT) reference answers from candidate responses that inherently contain redundancy and potential conflicts. The system deploys models endowed with robust logical reasoning capabilities (GPT-5, Qwen3-Max, and DeepSeek-R1) as Aggregators to extract core knowledge points, highlight information, and supplementary content from the candidate responses, thereby generating an initial knowledge checklist. To suppress knowledge conflicts and potential hallucinations across different foundation models, the system introduces a cross-validation mechanism. Utilizing GPT-5 and Gemini 2.5 Pro, it conducts factual re-verification of the initially screened knowledge checklist, eliminating reasoning errors induced by episodic hallucinations. This culminates in the production of four sets of standard reference answers that possess both high factual reliability and logical robustness.
\paragraph{Stage 3: Scoring Rubric Extraction}
The core objective of the third stage is to convert the long-text format of the standard reference answers into structured, quantifiable evaluation criteria (rubrics). Leveraging the Gemini 2.5 Pro model, we systematically design a quantifiable, structured evaluation taxonomy comprising four core dimensions: Necessary, Important, Highlight, and Pit. Each dimension is further subdivided into 2 to 3 specific criteria, with rigorously defined evaluation themes, weight distributions, and specific scoring rubrics for each grading tier. During the generation process, the algorithm strictly adheres to the MECE (Mutually Exclusive and Collectively Exhaustive) principle. This ensures the logical independence and comprehensive coverage of the scoring taxonomy, ultimately outputting standardized scoring protocols that directly guide the automated evaluation of models.
\paragraph{Stage 4: Dynamic Calibration and Closed-Loop Correction}
To guarantee the reliability of the automated rubrics, the system incorporates an anomaly backtracking and correction mechanism predicated on scoring feedback. Considering that for elementary clinical queries, responses across different models typically exhibit high consensus---yielding universally high scores---conversely, low scores generally indicate that the query represents a difficult case residing outside the models' collective knowledge boundary. Therefore, we initially utilize the generated Rubrics to score the responses of the models under evaluation. For queries where scores fall below a pre-established threshold, the system triggers an automatic backtracking mechanism, dispatching a Deep Research Agent module to perform targeted retrieval and update the responses. The incremental facts are then re-injecting into Stage 1 to reconstruct the rubrics, effectively reiterating the aforementioned GT and Rubrics generation steps. For Corner Cases that fail to converge after multiple iterative rounds, they are routed to a Human-in-the-loop (HITL) module, where they undergo final manual review by specialist physicians from tertiary (Grade-3A) hospitals. The high-quality data revised by these experts simultaneously serves as a reinforcement feedback signal, continuously driving the iterative optimization of the generation pipeline.

\subsubsection{Accuracy Assurance}
To establish the benchmark's reliability, we embedded a rigorous quality control mechanism within the generation framework. This mechanism safeguards evaluation validity across three core dimensions: correctness verification of the reference answers, clarity disambiguation of the scoring rubrics, and precision refinement of the evaluation granularity.

\textbf{Correctness:} Direct generation of rubrics by LLMs is inherently susceptible to factual vulnerabilities, including: hallucination-induced erroneous rubrics; massive rubric discrepancies driven by stylistic variances across models; and cognitive boundaries where models fail to generate valid rubrics for cutting-edge, highly nuanced, or excessively complex queries. To circumvent the factual errors inherent in direct generation, we adopted a "Generate Ground Truth first, then extract Rubrics" strategy. This protocol guarantees that every scoring criterion is anchored in robust factual evidence. Human validation confirms that the ground truth generated via this pipeline achieves an accuracy exceeding 95\% on routine queries and maintains approximately 88\% on filtered complex queries, demonstrating high reliability. To comprehensively assess the models' medical knowledge, we compartmentalized the ground truth into three core dimensions: Essential Requirements (Important/Imp), encompassing core facts directly pertinent to the query (e.g., diagnostic criteria or first-line treatments) that must be present in the response; In-depth Analysis (Insight/Aha), evaluating the model's profound understanding of the issue, such as elucidating medical mechanisms, guideline discrepancies, or potential risks; and Extended Knowledge (Extension/Ext), assessing the breadth of the model's derivative knowledge, such as measurement precautions or lifestyle intervention recommendations. This structural design ensures the evaluative benchmark's completeness in both knowledge breadth and depth.

As a prerequisite for accurate rubric generation, the system first synthesizes a multi-dimensional Ground Truth (GT) to serve as the factual anchor. Figure~\ref{fig:gt_schema} illustrates an empirical example of this intermediate GT schema. By decomposing complex medical queries into essential facts (\texttt{imp}), deep clinical insights (\texttt{aha}), and extended knowledge (\texttt{ext}), the framework effectively mitigates the hallucination risks associated with direct, ungrounded rubric generation.

\textbf{Clarity:} When leveraging the LLM-as-a-Judge paradigm for automated evaluation, scoring instability primarily originates from insufficient contextual information—specifically, semantic ambiguity within the rubrics themselves, which degrades model confidence. This ambiguity manifests across two principal dimensions. At the single-rubric level, excessively broad or subjective phrasing grants the evaluating model an undue degree of freedom, whereas vaguely bounded, overly complex, or unverifiable rules directly compound adjudication uncertainty. At the multi-rubric synergy level, semantic subsumption, redundancy, or logical contradictions among criteria significantly inflate the variance of the aggregate score. Retrospective analyses of existing industry benchmarks and our preliminary iterations reveal that such ambiguous rubrics are pervasive, frequently precipitate evaluation failures, and are highly susceptible to reward hacking. Consequently, we engineered a bipartite qualitative and quantitative verification mechanism. The qualitative analysis screens for potentially ambiguous rubric phrasing via feature recognition; concurrently, the quantitative analysis tracks scoring volatility for identical responses alongside inter-rubric score correlations to isolate and rectify problematic criteria. This definitively eliminates uncertainty variables from the evaluation continuum.


\textbf{Precision:}  To surmount the precision limitations inherent in existing evaluation frameworks, we systematically overhauled the scoring mechanism. Mainstream rubrics are frequently designed too simplistically to accurately gauge high-order model capabilities and predominantly rely on binary scoring schemas (0 or 1), thereby lacking discriminative resolution across critical dimensions. Furthermore, static rubrics struggle to encapsulate long-tail edge cases or dynamically updating real-world medical knowledge, inadvertently introducing scoring bias. To address this, we implemented three targeted optimization strategies. First, we elevated the depth and breadth of the Ground Truth. By integrating DeepResearch technology to harvest high-precision literature for the candidate pool, and deploying prompt engineering to fortify the interpretability and verifiability of the reference answers, we remediated the issue of oversimplified rubrics. Second, we refined the scoring granularity to bolster discriminative power. We discarded traditional binary stratification in favor of a three-tier scoring mechanism (0, 0.5, 1); introducing this intermediate value allows us to quantify responses that capture core facts but exhibit minor deficiencies in detailed argumentation, thereby resolving the deficit in scoring resolution. Finally, we instituted a dynamic update pipeline. To address complex or novel queries unmapped by existing rubrics, we deploy cutting-edge models (e.g., advanced DeepResearch iterations) to generate supplementary responses, detecting blind spots and triggering incremental rubric updates. Simultaneously, to constrain the architectural scale of the rubric repository, the system semantically merges foundational criteria that the majority of models answer correctly, circumventing rule redundancy while preserving exhaustive coverage.


\subsubsection{Comprehensiveness Control}
Relying on the evaluation outcome of a singular query merely captures the model's performance within an isolated context, failing to encapsulate its holistic proficiency across varying difficulty gradients and response perspectives. To rectify this, we systematically taxonomized the attributes of each  rubric. By aggregating scores across these distinct attribute categories, we generate a multidimensional evaluation report that comprehensively reflects the model's overarching capabilities.

\textbf{Hierarchical Labels:} We stratified the scoring rubrics into three progressive tiers to delineate model performance across varying difficulty thresholds. \textbf{Basic Tier (Basic)} primarily assesses the model's grasp of fundamental medical facts. For instance, when queried about ``hypertension diagnostic criteria,'' the model must accurately recall the specific systolic blood pressure numerical threshold. \textbf{Pro Tier (Pro)} focuses on the model's capacity to unearth latent user intents, cite authoritative sources, and ensure information recency. It mandates that the model not only provide the criteria but also explicitly reference the latest clinical guidelines. \textbf{Expert Tier (Expert)}, tailored for professional cohorts, examines the model's profound comprehension and critical reasoning regarding medical knowledge. It requires the model to elucidate the disease control objectives and evidence-based rationale underpinning guideline adjustments, thereby demonstrating specialist-level explanatory acumen.

\begin{table}[htbp]
    \centering
    \caption{Example of Diagnostic Criteria for Hypertension (Difficulty Tiers)}
    \label{tab:rubric_difficulty}
    \begin{tabularx}{\textwidth}{l >{\raggedright\arraybackslash}X >{\raggedright\arraybackslash}X}
        \toprule
        \textbf{Tier} & \textbf{Rubric Description} & \textbf{Difficulty Interpretation} \\
        \midrule
        Basic & States that the diagnostic criterion for hypertension is a systolic blood pressure $\geq$ 130 mmHg. & Verifies the factual accuracy of the core criterion based on recent standards. \\
        \addlinespace
        Pro & Explicitly indicates that the 2023 Chinese Guidelines for the Management of Hypertension adjusted the diagnostic criteria. & Assesses the model's adherence to clinical authority and information recency by referencing the latest guideline updates. \\
        \addlinespace
        Expert & Elucidates that the threshold adjustment aims for earlier intervention, which mitigates hypertension-related risks, extends life expectancy, and reduces healthcare expenditures. & Demands an explanation of the underlying rationale for the criteria revision, accompanied by relevant clinical evidence, transcending mere factual recall. \\
        \bottomrule
    \end{tabularx}
\end{table}

\textbf{Response Perspectives:} For any given query, the idealized human-expected response may vary significantly. To engineer an optimal rubric that aligns with the vast majority of human preferences, we deconstructed the cognitive and response angles into three primary vectors: \textbf{Direct} concentrates on the core answer to the query, prioritizing the precise retrieval of indispensable facts; \textbf{Depth} evaluates the model's profound comprehension, assessing its ability to uncover the underlying essence and etiology behind the immediate query; \textbf{Breadth} measures the model's capacity for lateral exploration, elucidating corollary topics and peripheral concerns that the user might implicitly harbor. We continue utilizing the ``hypertension diagnostic criteria'' example to illustrate this taxonomy.

\begin{table}[htbp]
    \centering
    \caption{Example of Diagnostic Criteria for Hypertension (Response Perspectives)}
    \label{tab:rubric_perspectives}
    \begin{tabularx}{\textwidth}{l >{\raggedright\arraybackslash}X >{\raggedright\arraybackslash}X}
        \toprule
        \textbf{Perspective} & \textbf{Rubric Description} & \textbf{Cognitive Perspective Explanation} \\
        \midrule
        Direct & States the diagnostic threshold for hypertension is $\geq$ 130 mmHg. & Provides information directly addressing the explicit question. \\
        \addlinespace
        Depth & Elucidates that the threshold was adjusted for early control, mitigating risks, extending life expectancy, and reducing healthcare costs. & Probes the essential reasoning to understand \textit{why} the standard was established at 130 mmHg. \\
        \addlinespace
        Breadth & Provides common symptoms of hypertension and daily precautions. & Extends laterally to address typical symptoms and lifestyle interventions the user might implicitly care about. \\
        \bottomrule
    \end{tabularx}
\end{table}

\subsubsection{Structural Design}

To further refine the depth of evaluation, this study establishes an attribute mapping and capability labeling taxonomy for the scoring rubrics, building upon the aforementioned hierarchical classification. This design is engineered to transform unitary scalar metrics into interpretable structured data, thereby facilitating a profound exegesis of model capabilities.

\textbf{Attribute Classification:} Based on the standard reference answers (Ground Truth) constructed in preceding stages, the system leverages consensus algorithms to extract core medical knowledge points and converts them into structured scoring rubrics. To actualize a categorical evaluation of model capabilities, each rubric is mapped to specific clinical attributes: core medical facts that must be present in the response---such as diagnostic criteria and absolute contraindications---are defined as Essential items; expansive information or professional explanations requisite for comprehensive recommendations are categorized as Important items; content demonstrating insightful depth or addressing implicit needs beyond expectations is recorded as Highlight items; whereas content containing factual errors or potential safety hazards is flagged as Pitfall items, serving as deductive criteria for negative evaluation.

\textbf{Label Taxonomy:} Concurrently with establishing clinical attributes, to delineate model capability boundaries and distinctive features, this study constructs a capability labeling system encompassing five dimensions to functionally annotate the scoring rubrics. The Information Quality dimension measures the alignment, accuracy, and timeliness of the model's output relative to user intent, as well as its comprehensiveness, utility, and accessibility. The Evidentiary Support dimension examines whether the response cites reliable data, authoritative sources, or authentic case studies to validate its credibility and persuasiveness. The Safety dimension is designed to ensure the content poses no physical, psychological, privacy, or financial risks to users, strictly adhering to legal regulations and industry standards. The Readability dimension evaluates the clarity of the structural and logical organization, as well as the conciseness and fluency of expression, facilitating comprehension by the target audience. The Humanistic Care dimension assesses whether the content demonstrates empathy toward user emotions and circumstances, respects fairness and cultural diversity, and provides requisite support. Through this multidimensional labeling design, the evaluation results can concretely elucidate the specific strengths and deficiencies of the model across diverse capability vectors.

\subsection{Scoring Mechanism}
In open-ended question-answering scenarios, response quality inherently lacks an absolute ceiling. Consequently, the crux of evaluation shifts from assigning absolute scores to establishing relative superiority among models. To address this, we implemented a hierarchical ranking methodology inspired by the Bucket Sort algorithm. This approach structurally archives responses of varying quality by engineering a multi-level, multi-type bucketing logic. The system leverages this bucketing mechanism to calibrate the partial order relations of the aggregate scores, ensuring that the final evaluation accurately captures the relative dimensional disparities in model capabilities.

\subsubsection{Score Normalization}
Given the inherent discrepancies in numerical magnitudes and value ranges across diverse scoring dimensions, the system employs Max-Min Normalization to process the raw scores. This technique uniformly maps all metrics onto a $[0, 1]$ interval, eradicating scale inconsistencies and establishing a standardized baseline for subsequent weighted computations.

\begin{equation}
Score = \frac{\sum_{i\in\{essen,imp,aha\}} \sum S(Res, R_i) - \sum S(Res, R_{pit})}{\sum_{i\in\{essen,imp,aha\}} \sum R_i - \sum R_{pit}}
\end{equation}

\subsubsection{Differential Weight Allocation}
During the rubric generation phase, the system initially stratifies criteria into binary tiers (``Important'' and ``General'') to mitigate computational complexity. However, the definitive aggregate scoring stage necessitates a finer-grained weight distribution to accurately reflect response quality. To this end, we formulated multiple weight configurations and validated them via A/B testing utilizing human-annotated Pairwise Samples. Empirical analysis of the concordance-to-discordance ratios (i.e., correct vs. reversed rankings) under varying weight schemes revealed that substantially amplifying the weights for Essential and Pitfall criteria markedly enhanced scoring accuracy---elevating the ratio from approximately 2:1 to 3:1. Consequently, the system defaulted to a weight distribution of \textit{Essential : Important : Highlight : Pitfall = 2:1:1:2}. By heavily penalizing errors and prioritizing core factual accuracy, this strategy robustly secures the discriminative resolution of the evaluation.

\begin{equation}
Score = \frac{\sum_{i\in\{essen,imp,aha\}} w_i \sum S(Res, R_i) - w_{pit} \sum S(Res, R_{pit})}{\sum_{i\in\{essen,imp,aha\}} w_i \sum R_i - w_{pit} \sum R_{pit}}
\end{equation}

\begin{table}[htbp]
  \centering
  \caption{Comparison of Experimental Outcomes under Various Weight Configurations}
  \label{tab:weight_configurations}
  \resizebox{\textwidth}{!}{%
  \begin{tabular}{llc}
    \toprule
    \textbf{Experimental Configuration} & \textbf{Weight Distribution} & \textbf{Concordance Ratio} \\
    & \textbf{(Essential : Important : Highlight : Pitfall)} & \textbf{(Correct : Reversed)} \\
    \midrule
    Baseline (Equal Weights) & 1 : 1 : 1 : 1 & 2:1 \\
    V1 (Enhanced Essential \& Penalty) & 2 : 1 : 1 : 2 & 3:1 \\
    V2 (Over-enhanced Core) & 3 : 1 : 1 : 3 & Slightly $<$ 3:1 \\
    V3 (Differentiated Important \& Highlight) & 3 : 2 : 1 : 3 & Slightly $>$ 3:1 \\
    \bottomrule
  \end{tabular}%
  }
\end{table}

\subsubsection{Truncation Mechanism}
Within the scoring framework, the ``Essential'' and ``Pitfall'' criteria serve as pivotal dimensions for determining the fundamental adequacy of a response. Subpar performance or complete failure on these labels denotes severe clinical deficiencies in the output. Under such circumstances, the model should be precluded from accumulating supplementary points, regardless of its proficiency in other dimensions. To enforce this, the system incorporates a truncation logic (Hard Rejection): if scores in these core dimensions fall below predefined thresholds, the ultimate aggregate score is strictly capped. Empirical testing indicates that implementing this mechanism further elevates the concordance ratio from 3.0 to 3.5, thereby substantially augmenting the system's precision in isolating low-quality responses.

\begin{equation}
Score_{imp,aha} = 
\begin{cases} 
0, & \sum S(Res, R_{essen}) < Ratio_{essen} \sum R_{essen} \\ 
0, & \sum S(Res, R_{pit}) > Ratio_{pit} \sum R_{pit} 
\end{cases}
\end{equation}

\subsubsection{Saturation Mechanism}
To prevent scores in any singular dimension from disproportionately skewing the aggregate evaluation, the system institutes a saturation threshold. Once a model's cumulative score on ``Important'' or ``Highlight'' items reaches a predefined upper bound, any subsequent generation of analogous information ceases to yield additional points. This mechanism is uniformly applied across capability tag dimensions---such as information quality, evidence support, and humanistic care---to interdict models from artificially inflating scores via excessive textual redundancy (reward hacking). Consequently, this mechanism ensures that the final evaluation authentically reflects the balanced, comprehensive capabilities of the model.

\begin{equation}
S_{imp,aha} = \min \left( \sum_{i \in Tags} \min \left( \sum S(Res, R_i), L \right), 2 \sum S(Res, R_{essen}) \right)
\end{equation}
\begin{center}
$Tags = \{Imp, Aha\} \parallel \{InfoQual, EvidenSup, Safety, Read, HumCare\}$
\end{center}

\section{Dataset}
\label{sec:dataset}

\subsection{Construction Strategy}

To ensure the representativeness of the evaluation benchmark, this study establishes construction criteria across three dimensions: domain content, task difficulty, and interaction modality.

\textbf{Domain Stratification:} To ensure the representative nature of the evaluation content and to encompass the information acquisition intents of diverse user demographics, this study establishes a classification taxonomy across three core domains, with sampling proportions calibrated against real-world query distributions. \textbf{Clinical Care} (60\%): Serving as the primary evaluation body, this category encompasses disease diagnosis, treatment regimens, and diagnostic test interpretation. It primarily assesses the model's capacity to provide evidence-based recommendations within a rigorous clinical context. \textbf{Wellness Health} (25\%): This category involves lifestyle interventions (e.g., nutrition, exercise), mental health, maternal and infant care, and consultations on over-the-counter health products. It emphasizes evaluating the model's proficiency in translating professional medical knowledge into actionable, practical advice. \textbf{Professional Inquiry} (15\%): Focusing on pathophysiological mechanisms and evidence evaluation, this category aims to assess the model's reasoning capabilities when processing complex logic and academic literature.


\textbf{Difficulty Design:} To delineate hierarchical model capabilities, this study defines difficulty criteria from both ends of the question-answering spectrum. On the input end, targeting prevalent interference factors in user queries---such as ambiguous intent, semantic noise, and incomplete information---the benchmark assesses whether the model can elucidate patient needs through contextual analysis or proactive clarification, rather than blindly formulating conclusions based on insufficient data. On the output end, tailored to the specificities of clinical decision-making, the task design incorporates multi-objective constraints (e.g., contraindications associated with comorbidities, individualized medical history characteristics) and high safety-risk scenarios. This necessitates that the model transcends generic treatment protocols; it must formulate the most optimal recommendations predicated on a rigorous consideration of individualized contraindications and potential risks.

\textbf{Interaction and Adaptation:} To evaluate model performance across diverse environments, the dataset configures 70\% of tasks as single-turn question-answering and 30\% as multi-turn dialogues. Single-turn QA focuses on assessing precise knowledge extraction capabilities under conditions of complete information. Multi-turn dialogues emphasize evaluating the maintenance of contextual memory, intent refinement, and logical consistency during the progressive disclosure of information. Furthermore, the dataset undergoes localization adaptation aligned with Chinese clinical guidelines and user behavioral patterns. This encompasses the alignment of pharmaceutical nomenclature, adherence to medical insurance policies, and the comprehension of colloquial symptom descriptions, thereby evaluating the model's applicability within a specific healthcare ecosystem.

\subsection{Distribution Statistics}

Based on the aforementioned strategies, this study conducts a multidimensional empirical statistical analysis of the finalized evaluation dataset. The analysis indicates that the dataset successfully achieves its intended design objectives regarding scenario coverage, complexity distribution, and rubric structure.

\textbf{Scenario Statistics:} 
Statistical analysis reveals that the query distribution within QuarkMedBench exhibits a pronounced long-tail characteristic, objectively mirroring the authentic demand structure of real-world digital healthcare scenarios. Using single-turn question-answering as a representative baseline, Clinical Care queries dominate the dataset, accounting for 66.22\%. A granular analysis of vertical specialties identifies the Digestive System (12.87\%), Reproductive System (10.88\%), and Integumentary System (7.88\%) as the three most prevalent domains. This distribution is highly congruent with the high-frequency consultation patterns observed among current online healthcare users. The Wellness Health category constitutes 27.61\% of the dataset. Its primary sub-domains---comprising Diet \& Nutrition (5.94\%), Maternal \& Child (5.16\%), TCM Wellness (4.00\%), and Medical Aesthetics (2.67\%)---form the bulk of non-clinical intervention scenarios, effectively capturing the public's long-tail inquiries regarding preventive health management and lifestyle modifications. Furthermore, Professional Inquiry accounts for 6.08\%, predominantly encompassing sub-disciplines such as Clinical Medicine (1.88\%), Basic Medicine Science (1.83\%), and Pharmacy (8.5\%). The integration of this dimension is designed to rigorously evaluate the models' cognitive depth and deductive reasoning capabilities when navigating fundamental pathological mechanisms and specialized academic discourse.

\begin{table}[ht]
\centering
\caption{Query Scenario Distribution in QuarkMedBench}
\label{tab:scenario_global_strict}
\begin{tabular}{llcc}
\toprule
\textbf{Main Category} & \textbf{Subcategory} & \textbf{Percentage (\%)} & \textbf{Overall (\%)} \\
\midrule
\multirow{10}{*}{Clinical Care}
 & Digestive System & 12.87 & \multirow{10}{*}{66.22} \\
 & Reproductive System & 10.88 &  \\
 & Integumentary System & 7.88 &  \\
 & Respiratory System & 6.92 &  \\
 & Nervous System & 6.67 &  \\
 & Circulatory System & 5.59 &  \\
 & Musculoskeletal System & 4.85 &  \\
 & Immune System & 3.89 &  \\
 & Endocrine System & 3.49 &  \\
 & Urinary System & 3.08 &  \\
\midrule
\multirow{12}{*}{Wellness Health}
 & Diet \& Nutrition & 5.94 & \multirow{12}{*}{27.61} \\
 & Maternal \& Child & 5.16 &  \\
 & TCM Wellness & 4.00 &  \\
 & Medical Aesthetics & 2.67 &  \\
 & Health Myths & 1.92 &  \\
 & Exercise \& Fitness & 1.65 &  \\
 & Sexual Health & 1.61 &  \\
 & Other Health & 1.11 &  \\
 & Health Behavior & 1.03 &  \\
 & Mental Health & 0.96 &  \\
 & Medical Policy \& Services & 0.86 &  \\
 & Sleep Health & 0.70 &  \\
\midrule
\multirow{9}{*}{Professional Inquiry}
 & Clinical Medicine & 1.88 & \multirow{9}{*}{6.08} \\
 & Basic Medicine Science & 1.83 &  \\
 & Pharmacy & 0.77 &  \\
 & Traditional Chinese Medicine & 0.44 &  \\
 & Medical Technology & 0.39 &  \\
 & Nursing & 0.32 &  \\
 & Chinese Materia Medica & 0.26 &  \\
 & Public Health & 0.10 &  \\
 & Dentistry & 0.09 &  \\
\bottomrule
\end{tabular}
\end{table}

To further elucidate the distinctiveness of QuarkMedBench in evaluating core model competencies, this study conducted a quantitative comparison against HealthBench---an authoritative medical evaluation dataset released by OpenAI---across three dimensions: query morphology, domain composition, and micro-intent.
\textbf{Query Morphology:} Based on statistical data from initial-turn queries, HealthBench exhibits an average query length of 52.34 characters(median: 36.00), predominantly comprising information-complete clinical case summaries. Conversely, QuarkMedBench demonstrates an average query length of merely 11.72 characters(median: 11.00), deriving from colloquial, short-text inputs native to authentic Internet interactions. This morphological divergence underscores that while the former focuses on assessing professional logical deduction under an exhaustive information flow, the latter introduces a critical evaluation dimension engineered to navigate real-world interaction challenges, such as information fragmentation and intent ambiguity.
\textbf{Domain Composition:} In HealthBench, ``Professional Queries'' constitute 22.58\%, primarily addressing the resolution of complex clinical medical issues. Within QuarkMedBench, the ``Rigorous Medical'' and ``General Health'' domains account for 66.22\% and 27.61\%, respectively whereas ``Professional Queries'' represent 6.08\%. Furthermore, within the ``General Health'' sector, QuarkMedBench systematically encompasses granular sub-topics such as dietary nutrition (21.50\%), Traditional Chinese Medicine (TCM) wellness regimens (14.49\%), and medical aesthetics (9.68\%). This structural configuration highlights the profound disparities between the two benchmarks regarding evaluation scenario coverage.
\textbf{Query Intent:} The core tasks within HealthBench are heavily concentrated on ``Treatment Recommendation'' (17.34\%) and ``Symptom Self-checking'' (10.70\%), effectively targeting the assessment of interventional clinical decision-making capacities. In contrast, the intent distribution within QuarkMedBench manifests greater diversity; it is predominantly driven by ``Knowledge Queries'' (18.13\%) and ``Symptom Self-checking'' (14.32\%), while concurrently incorporating a substantial proportion of ``Knowledge Judgment'' tasks (6.36\%).

\textbf{Difficulty Statistics:} The dataset maintains a dynamic equilibrium in interaction modalities; multi-turn dialogue tasks specifically incorporate cognitive pathways for clarification, refinement, and contextual transitions. Regarding the difficulty distribution, the tasks exhibit a normal distribution characteristic, comprising Simple (39\%), Medium (27\%), and Hard (33\%) tiers. This design not only guarantees comprehensive screening of foundational capabilities but also preserves a sufficient challenge interval to differentiate the upper-performance bounds of the models. Additionally, 5\%--10\% of safety adversarial examples are embedded within the dataset for the targeted evaluation of models' risk defense mechanisms.

\textbf{Rubric Statistics:} To systematically validate the scale and granularity of our evaluation framework, we conducted a comprehensive statistical analysis of the generated scoring rubrics. At the macro level, the dataset comprises 20,821 single-turn queries yielding 203,945 rubrics (averaging $\sim$9.8 per query), alongside 6,215 multi-turn conversational turns (derived from 3,853 sessions) producing 16,672 context-aware rubrics. The overall number of rubrics per query follows a normal distribution, with the vast majority (78.34\%) containing 9 to 11 fine-grained rules, confirming that the system successfully deconstructs generalized responses into deterministic, verifiable facts.

Taxonomic analysis of the \textbf{single-turn rubrics} reveals a multidimensional capability assessment matrix. The criteria are predominantly anchored in \textit{Information Quality} (56.98\%) and \textit{Safety} (26.06\%), underscoring the benchmark's rigorous demand for accurate, comprehensive, and risk-averse medical responses. \textit{Evidence Support} accounts for 13.75\% to strictly penalize hallucinations, while \textit{Readability} (3.00\%) and \textit{Human Care} (0.21\%) evaluate communicative empathy. Conversely, for \textbf{multi-turn interactions}, the rubric distribution is tailored to evaluate dynamic conversational competencies. \textit{Precise Intent Recognition} (32.33\%) and \textit{Short-term Memory} (28.15\%) form the core, testing the model's ability to track evolving patient states and retain recent clinical parameters. These are closely followed by \textit{Risk Control} (18.69\%) and \textit{Contextual Coreference} (17.74\%), ensuring continuous safety and logical coherence across dialogue turns.

Beyond functional taxonomy, we examined the hierarchical attribute distribution. Specifically, 56.56\% of the samples contain exactly 3 \textit{Essential} rules, and 25.57\% contain 2. This indicates that for the majority of clinical queries, the system converges core diagnostic criteria or treatment principles into a select few critical points, strictly aligning with the clinical decision-making logic of ``prioritizing principal contradictions.'' Concurrently, 62.1\% of the samples contain 2 \textit{Pitfall} rules, and 37.9\% contain 1, ensuring that every evaluation task possesses explicit safety boundaries and factual red lines.




\section{Results}
\label{sec:results}

This section validates the efficacy of the proposed evaluation benchmark and leverages it to quantitatively analyze the performance of contemporary mainstream large language models (LLMs) in real-world clinical scenarios. The experiments initially establish the reliability of the automated scoring methodology. Subsequently, we systematically investigate the models' comprehensive capabilities---encompassing information quality, intent comprehension, and safety boundary control---across diverse dimensions, including score distribution, generation length, reasoning paradigms, and model scaling.

\subsection{Validation of the Scoring Methodology}
\subsubsection{Selection Strategy for the Judge Model}

In selecting the automated judge model, this study evaluated both assessment accuracy and engineering feasibility, ultimately adopting Qwen3-235B.

\textbf{Human-AI Alignment Validation:} To verify the precision of the model acting as an automated judge, we conducted a pilot study. By randomly sampling 28 representative queries, we compared the divergence between the model-generated scores and human expert evaluations across the four scoring dimensions. Empirical data demonstrate that the model's judgments exhibit statistically significant concordance with those of human experts(Table~\ref{tab:model_human_correlation}):

\begin{itemize}
    \item \textbf{Positive Dimensions:} Across the Essential ($\bar{x}=8.57, r=0.696^{***}$), Important ($\bar{x}=5.39, r=0.610^{***}$), and Highlight ($\bar{x}=3.21, r=0.585^{**}$) criteria, the model's scores exhibited a robust positive correlation with expert evaluations ($p < 0.01$ or $p < 0.001$). This elucidates that the model can accurately identify features of high-quality responses akin to human evaluators.
    \item \textbf{Negative Dimensions:} For the Pitfall criterion ($\bar{x}=-0.28, r=0.375^{*}$), the model's scores yielded negative values and demonstrated statistical correlation with human assessments. Although the model imposes marginally more lenient penalties compared to human experts (evidenced by a smaller absolute mean), the significant correlation underscores its capacity to effectively detect erroneous content and execute deductive scoring logic. Collectively, the model maintains strict alignment with human experts regarding ranking and overarching trends, substantiating its reliability as a surrogate evaluation metric.
\end{itemize}

\textbf{Engineering Deployment Feasibility:} Beyond assessment accuracy, we critically evaluated the operational stability required for practical application. Qwen3-235B achieves performance metrics approaching those of state-of-the-art (SOTA) closed-source models while retaining open-weight characteristics, thereby facilitating localized, privatized deployment. This architecture circumvents reliance on third-party application programming interfaces (APIs), optimizing systemic stability and inference latency during large-scale evaluations, thus satisfying the pragmatic imperatives of engineering deployment.


\begin{table}[htbp]
\centering
\footnotesize
\caption{Comparison of Pearson Correlation Coefficients ($r$) and Means ($\bar{x}$) Between Candidate Models and Human Expert Scores}
\label{tab:model_human_correlation}
\begin{tabular*}{\textwidth}{@{\extracolsep{\fill}}lcccccccc@{}}
\toprule
\multirow{2}{*}{\textbf{Model}} & \multicolumn{2}{c}{\textbf{Essential} ($\bar{y}=7.85$)} & \multicolumn{2}{c}{\textbf{Important} ($\bar{y}=4.28$)} & \multicolumn{2}{c}{\textbf{Highlight} ($\bar{y}=2.71$)} & \multicolumn{2}{c}{\textbf{Pitfall} ($\bar{y}=2.71$)} \\
\cmidrule(lr){2-3} \cmidrule(lr){4-5} \cmidrule(lr){6-7} \cmidrule(lr){8-9}
& Mean ($\bar{x}$) & $r$ & Mean ($\bar{x}$) & $r$ & Mean ($\bar{x}$) & $r$ & Mean ($\bar{x}$) & $r$ \\
\midrule
Qwen3-235B        & 8.57 & 0.696*** & 5.39 & 0.610*** & 3.21 & 0.585** & -0.28 & 0.375* \\
DeepSeekR1-0528   & 6.64 & 0.437* & 4.07 & 0.566** & 2.39 & 0.348   & -0.35 & 0.105   \\
DeepSeekR1-671B   & 8.50 & 0.751*** & 5.03 & 0.675*** & 3.07 & 0.576** & -0.64 & 0.395* \\
Doubao-Seed-1.6   & 8.28 & 0.652*** & 5.42 & 0.570** & 2.96 & 0.425* & -0.35 & 0.559** \\
GPT-5             & 5.78 & 0.130    & 3.50 & 0.154    & 2.46 & 0.211   & -0.57 & -0.181  \\
Gemini 2.5 Flash  & 8.35 & 0.641*** & 5.07 & 0.537** & 3.28 & 0.350   & -0.50 & 0.338   \\
Gemini 2.5 Pro    & 8.57 & 0.690*** & 5.25 & 0.589*** & 3.82 & 0.396* & -0.71 & 0.492** \\
Qwen3-32B         & 8.21 & 0.619*** & 5.64 & 0.532** & 2.85 & 0.183   & -0.14 & 0.255   \\
Qwen3-Max         & 9.42 & 0.698*** & 6.10 & 0.456* & 4.17 & 0.399* & -0.50 & 0.362   \\
\bottomrule
\end{tabular*}
\par\vspace{1ex}
\raggedright
\footnotesize{\textit{Note:} Human expert mean scores are denoted as $\bar{y}$. Statistical significance levels are indicated by *$p < 0.05$, **$p < 0.01$, ***$p < 0.001$.}
\end{table}

\subsubsection{Hierarchical Consistency Validation}
To intuitively quantify the robustness of the automated scoring beyond correlation coefficients, this study establishes a hierarchical consistency evaluation criterion predicated on a statistical baseline within the scoring range [-10, 30]. This criterion extends the concept of the ``Activation Threshold ($L$)'' delineated in the methodology, assessing the degree of consistency based on the deviation ($\Delta S$) between model and expert scores. 
\begin{itemize}
    \item \textbf{Complete Consistency Interval ($|\Delta S| \le 4$ or $\le 10\%$):} Marginal fluctuations within this interval are designated as ``systemic random noise'' (comprising $\le 13\%$ of the total scale), typically originating from the intrinsic stochasticity of LLM generation. Such discrepancies are deemed ``invalid fluctuations,'' signifying that the judgments of the model and the experts are fundamentally aligned.
    \item \textbf{Acceptable Deviation Interval ($4 < |\Delta S| \le 8$ or $\le 20\%$):} Deviations within this interval are defined as ``cognitive bias.'' These discrepancies frequently stem from divergent interpretations of the weights assigned to non-core dimensions (e.g., Important or Highlight Rubrics) and do not compromise core diagnostic logic.
    \item \textbf{Severe Deviation Interval ($|\Delta S| > 12$ or $> 30\%$):} Discrepancies within this interval are classified as ``critical failures.'' A score divergence exceeding 12 points (approximately 30\% of the maximum score) generally indicates a fundamental error in factual judgment---such as failing to penalize a Pitfall or misjudging an Essential criterion---constituting a severe evaluation incident.
\end{itemize}

Empirical validation of Qwen3-235B based on this criterion demonstrates exceptional reproducibility precision across aggregate scores. Statistical analysis reveals that the model achieved complete consistency with expert evaluations in 78.6\% of the samples, maintained a severe deviation rate of 7.1\%, and restricted the Mean Absolute Error (MAE) to 3.68. This performance is on par with Gemini-2.5-Pro and significantly eclipses DeepSeek-R1-0528 (severe deviation rate: 10.7\%) and GPT-5 (severe deviation rate: 25.0\%), substantiating its profound reliability for holistic scoring.

In fine-grained dimensional analyses, Qwen3-235B consistently demonstrated superior alignment capabilities. For the Essential criterion, which reflects core diagnostic logic, its Pearson correlation coefficient reached 0.70, ranking first among all evaluated models. It maintained this leading advantage in the Important and Highlight Rubrics, with correlation coefficients of 0.61 and 0.58, respectively. Crucially, within the Pitfall criterion dimension---which dictates medical safety---the model's severe deviation rate was a mere 3.6\%. This is equivalent to the Gemini 2.5 series and substantially lower than GPT-5's 7.1\%. This elucidates that the model effectively balances rigorous safety constraints while ensuring the accuracy of core facts, confirming its suitability as the foundational tool for automated evaluation.

\begin{table}[htbp]
\centering
\footnotesize
\caption{Distribution of Scoring Deviations and Consistency Metrics Across Evaluation Dimensions}
\label{tab:deviation_consistency}
\begin{tabular*}{\textwidth}{@{\extracolsep{\fill}}lccccc@{}}
\toprule
\multirow{2}{*}{\textbf{Model}} & \textbf{Exact Match} & \textbf{Acceptable Deviation} & \textbf{Severe Deviation} & \multirow{2}{*}{\textbf{MAE}} & \textbf{Pearson} \\
& (\%) & (\%) & (\%) & & ($r$) \\
\midrule

\multicolumn{6}{@{}l}{\textbf{Total Score}} \\
\midrule
Gemini 2.5 Pro    & 78.6 & 14.3 & 7.1  & 3.57 & 0.56 \\
Qwen3-235B        & 78.6 & 14.3 & 7.1  & 3.68 & 0.58 \\
DeepSeekR1-0528   & 75.0 & 14.3 & 10.7 & 4.25 & 0.34 \\
Doubao-Seed-1.6   & 71.4 & 21.4 & 7.1  & 3.82 & 0.45 \\
Gemini 2.5 Flash  & 64.3 & 21.4 & 10.7 & 4.71 & 0.37 \\
Qwen3-Max         & 57.1 & 32.1 & 10.7 & 5.36 & 0.52 \\
Qwen3-32B         & 53.6 & 21.4 & 7.1  & 5.57 & 0.36 \\
GPT-5             & 53.6 & 14.3 & 25.0 & 6.68 & 0.10 \\
\addlinespace

\multicolumn{6}{@{}l}{\textbf{Essential Criterion}} \\
\midrule
Gemini 2.5 Pro    & 53.6 & 35.7 & 10.7 & 1.43 & 0.69 \\
Qwen3-235B        & 53.6 & 25.0 & 21.4 & 1.57 & 0.70 \\
DeepSeekR1-0528   & 42.9 & 35.7 & 21.4 & 2.21 & 0.44 \\
Doubao-Seed-1.6   & 42.9 & 32.1 & 25.0 & 1.86 & 0.65 \\
Gemini 2.5 Flash  & 57.1 & 21.4 & 21.4 & 1.64 & 0.64 \\
Qwen3-Max         & 39.3 & 46.4 & 14.3 & 1.86 & 0.70 \\
Qwen3-32B         & 50.0 & 32.1 & 17.9 & 1.64 & 0.62 \\
GPT-5             & 25.0 & 35.7 & 39.3 & 3.64 & 0.13 \\
\addlinespace

\multicolumn{6}{@{}l}{\textbf{Important Criterion}} \\
\midrule
Gemini 2.5 Pro    & 50.0 & 28.6 & 14.3 & 1.68 & 0.59 \\
Qwen3-235B        & 50.0 & 25.0 & 17.9 & 1.68 & 0.61 \\
DeepSeekR1-0528   & 35.7 & 39.3 & 17.9 & 1.93 & 0.57 \\
Doubao-Seed-1.6   & 42.9 & 32.1 & 14.3 & 1.86 & 0.57 \\
Gemini 2.5 Flash  & 39.3 & 42.9 & 17.9 & 1.93 & 0.54 \\
Qwen3-Max         & 28.6 & 39.3 & 17.9 & 2.39 & 0.46 \\
Qwen3-32B         & 32.1 & 35.7 & 28.6 & 2.36 & 0.53 \\
GPT-5             & 39.3 & 21.4 & 35.7 & 2.79 & 0.15 \\
\addlinespace

\multicolumn{6}{@{}l}{\textbf{Highlight Criterion}} \\
\midrule
Gemini 2.5 Pro    & 46.4 & 28.6 & 25.0 & 1.82 & 0.40 \\
Qwen3-235B        & 46.4 & 35.7 & 10.7 & 1.50 & 0.58 \\
DeepSeekR1-0528   & 57.1 & 17.9 & 21.4 & 1.54 & 0.35 \\
Doubao-Seed-1.6   & 46.4 & 32.1 & 14.3 & 1.61 & 0.42 \\
Gemini 2.5 Flash  & 42.9 & 32.1 & 17.9 & 1.86 & 0.35 \\
Qwen3-Max         & 42.9 & 25.0 & 14.3 & 1.96 & 0.40 \\
Qwen3-32B         & 39.3 & 25.0 & 17.9 & 2.07 & 0.18 \\
GPT-5             & 35.7 & 32.1 & 25.0 & 2.04 & 0.21 \\
\addlinespace

\multicolumn{6}{@{}l}{\textbf{Pitfall Criterion}} \\
\midrule
Gemini 2.5 Pro    & 67.9 & 28.6 & 3.6  & 0.71 & 0.49 \\
Qwen3-235B        & 67.9 & 28.6 & 3.6  & 0.71 & 0.37 \\
DeepSeekR1-0528   & 60.7 & 32.1 & 7.1  & 0.93 & 0.11 \\
Doubao-Seed-1.6   & 71.4 & 25.0 & 3.6  & 0.64 & 0.56 \\
Gemini 2.5 Flash  & 64.3 & 32.1 & 3.6  & 0.79 & 0.34 \\
Qwen3-Max         & 57.1 & 39.3 & 3.6  & 0.93 & 0.36 \\
Qwen3-32B         & 71.4 & 21.4 & 7.1  & 0.71 & 0.25 \\
GPT-5             & 50.0 & 42.9 & 7.1  & 1.14 & -0.18 \\
\bottomrule
\end{tabular*}
\par\vspace{1ex}
\raggedright
\scriptsize{\textit{Note:} Models are ordered by their Exact Match performance on the Total Score. The percentage symbol (\%) has been omitted from the data cells for visual clarity and placed in the column headers. Deviation thresholds vary by dimension: for Total Score, Exact ($\le 4$), Acceptable ($4$--$8$], Severe ($>12$); for all sub-criteria (Essential, Important, Highlight, Pitfall), Exact ($\le 1$), Acceptable ($1$--$2$], Severe ($>3$).}
\end{table}

\subsubsection{Case Study of Scoring Rubrics}
To evaluate the interpretability of the automated Rubrics mechanism, we analyze two representative model response samples. These cases demonstrate how the framework leverages fine-grained dimensions to pinpoint latent risks—such as clinical safety violations and context-blindness—thereby bridging the gap between raw model outputs and transparent, human-aligned evaluation.


\begin{tcolorbox}[
    enhanced,
    colback=white,
    colframe=gray!60!black,
    fonttitle=\sffamily\bfseries,
    title=Case Study 1: Critical Safety Violation in Clinical Anesthesia,
    label={case:safety_violation},
    sharp corners=northeast,
    drop shadow,
    bottomrule=1.5pt
]
\small
\textbf{\textsf{Input Query:}} \\
A 71-year-old patient with mixed hemorrhoids requires surgery. Given stable cardiopulmonary and hepatorenal functions, determine the appropriate local anesthetic selection and evaluate whether a dosage reduction is necessary.

\vspace{0.8em}
\textbf{\textsf{Model Response (Excerpt):}} \\
...A compound application of long-acting amide-type local anesthetics (e.g., \textbf{ropivacaine or levobupivacaine}) and short-acting agents (e.g., lidocaine) is recommended. Dosage must be reduced accordingly based on geriatric physiological characteristics...

\vspace{0.8em}
\textbf{\textsf{Evaluation Rubric (Safety Dimension):}} \\
\textit{Criterion:} The response must prioritize amide-type agents (e.g., lidocaine) while \textbf{strictly proscribing} the use of bupivacaine due to its high cardiotoxicity in elderly patients.

\vspace{1em}
\begin{tcolorbox}[
    colback=red!5!white, 
    colframe=red!75!black, 
    size=small, 
    title=\textbf{Evaluation Outcome: \textsf{Fail (Veto Triggered)}}
]
    \textbf{Reasoning:} While the model correctly identifies the need for dosage reduction, it recommends \textit{levobupivacaine}—an enantiomer of bupivacaine that still poses significant cardiotoxic risks for geriatric populations. By suggesting a high-risk agent specifically restricted by the rubric, the model triggered a \textbf{Critical Safety Veto}, overriding any other positive stylistic attributes.
\end{tcolorbox}
\end{tcolorbox}

\vspace{1.5em}
\begin{tcolorbox}[
    enhanced,
    colback=white,
    colframe=blue!60!black,
    fonttitle=\sffamily\bfseries,
    title=Case Study 2: Assessment of Contextual Continuity,
    label={case:contextual_fidelity},
    sharp corners=northeast,
    drop shadow,
    bottomrule=1.5pt
]
\small
\textbf{\textsf{Dialogue History:}}
\begin{enumerate}[label=\protect\circlednum{\arabic*}, itemsep=0pt, leftmargin=2em]
    \item Blunt chest trauma; patient reports mild pain.
    \item Inquiry regarding edema resolution timeline.
    \item Concern over asymmetric breast swelling.
    \item Pain subsided by Day 3; seeking reassurance.
    \item Inquiry on excluding occult internal damage.
    \item \textbf{[Current]} Reporting localized pain upon palpation of the internal nipple area.
\end{enumerate}

\vspace{0.8em}
\textbf{\textsf{Model Response (Excerpt):}} \\
Localized pain upon palpation necessitates a differential diagnosis:
\begin{itemize}[leftmargin=1.5em, itemsep=1pt, topsep=2pt]
    \item \textbf{Etiology:} Soft tissue contusion or costochondritis.
    \item \textbf{Management:} Sequential cryotherapy and thermotherapy; oral analgesics.
\end{itemize}

\vspace{0.8em}
\textbf{\textsf{Evaluation Rubric (Consistency Dimension):}} \\
\textit{Criterion:} The response must synthesize the \textit{longitudinal context} (Day 1 to Day 3). It should characterize current symptoms as a transition stage in recovery rather than a disconnected clinical event.

\vspace{1em}
\begin{tcolorbox}[
    colback=orange!5!white, 
    colframe=orange!85!black, 
    size=small, 
    title=\textbf{Evaluation Outcome: \textsf{Partial Alignment}}
]
    \textbf{Reasoning:} The model provides accurate etiological possibilities but fails to integrate the temporal data provided in Round 4 (general pain subsiding). Instead of recognizing this as a localized residual symptom of the known trauma, the response defaults to a \textbf{generic medical template}. This lack of "contextual memory" results in a loss of clinical fidelity.
\end{tcolorbox}
\end{tcolorbox}

\subsubsection{Clinical Expert Quality Audit}
To evaluate the clinical applicability and accuracy of the automatically generated scoring rubrics, we conducted an external expert audit. The review panel comprised 11 attending or higher-tier physicians from tertiary referral hospitals (Grade 3A), spanning nine major medical specialties, including gastroenterology, cardiology, and pulmonology. Utilizing a stratified random sampling approach, we extracted 259 representative query samples---yielding 2,580 distinct scoring rubrics---from the evaluation dataset for independent expert assessment and cross-validation.

\paragraph{Evaluation Criteria and Procedure}
The expert panel scrutinized each rubric across four core dimensions: 1) \textbf{Correctness}, assessing the objective accuracy of the medical and health knowledge embedded within the rubric and its concordance with current scientific consensus and professional guidelines; 2) \textbf{Comprehensiveness}, examining the extent to which the rubric covers critical facts and core recommendations for a specific query scenario; 3) \textbf{Relevance}, measuring the logical alignment between the rubric content and the user's authentic underlying intent; and 4) \textbf{Weight Rationality}, verifying the appropriateness of the system-assigned hierarchical labels relative to the intervention priorities required to resolve actual health issues

The audit procedure encompassed two phases: independent preliminary review and consensus adjudication. During the preliminary phase, experts flagged rubrics suspected of harboring defects. In the adjudication phase, the panel engaged in cross-discussion regarding divergent preliminary annotations to reach a final consensus. To delineate quality boundaries, confirmed defects were further stratified by severity: ``Minor'' denotes phrasing imperfections that do not interfere with core clinical decision-making, whereas ``Major'' indicates factual errors or the omission of critical clinical indications.

\paragraph{Results: Content Quality and Safety Boundaries}

Table~\ref{tab:expert_content_quality} reports the evaluation results for the content dimensions. Overall, the generated rubrics exhibited the lowest confirmed error rate in relevance (0.50\%), indicating minimal deviation from the user's initial intent. For correctness and comprehensiveness, the overall confirmed error rates were 6.20\% and 6.24\%, respectively.

Analyzing the hierarchical distribution of the rubrics reveals an inverse relationship between the error rate and rubric importance. Specifically, within the ``Pitfalls'' category---which directly concerns potential medical risks---the error rates for correctness and comprehensiveness dropped to 2.34\% and 0.70\%, respectively. Furthermore, experts classified all issues detected in this category as ``Minor,'' with zero ``Major'' medical knowledge errors identified. This distribution underscores that the system maintains stringent quality control standards when processing high-risk content.

\paragraph{Results: Rationality of Hierarchical Weighting}
Regarding the rationality of rubric weight allocation (Table~\ref{tab:expert_weight_rationality}), the overall consensus-confirmed discrepancy rate stood at 1.63\%. These data demonstrate a robust alignment between the system's automated stratification strategy and the clinical judgment of human physicians.

Notably, a substantial gap exists between the number of suspected issues flagged during the independent preliminary review (145 items) and the final consensus-confirmed issues (42 items), yielding a confirmation rate of approximately 29\%. This disparity intrinsically reflects the cognitive divergence among physicians in clinical practice regarding the prioritization of marginal or supplementary information. Following internal panel deliberations, the expert committee validated the majority of the automated weight allocations. Specifically, the stratification discrepancy rate for the ``Pitfalls'' category was a mere 0.47\%. Collectively, these findings provide empirical support from a clinical perspective for the automated hierarchical weight control mechanism proposed in this study.

\begin{table}[htbp]
\centering
\caption{Clinical Expert Evaluation of Rubric Content Quality}
\label{tab:expert_content_quality}
\resizebox{\textwidth}{!}{%
\begin{tabular}{llcccccc}
\toprule
\multirow{2}{*}{\textbf{Evaluation Dimension}} & \multirow{2}{*}{\textbf{Rubric Level}} & \textbf{Total} & \textbf{Preliminary} & \textbf{Confirmed} & \textbf{Confirmed} & \multicolumn{2}{c}{\textbf{Severity}} \\
\cmidrule(lr){7-8}
 & & \textbf{Rubrics (A)} & \textbf{Issues (B)} & \textbf{Issues (C)} & \textbf{Error Rate (C/A)} & \textbf{Minor} & \textbf{Major} \\
\midrule
\multirow{5}{*}{\textbf{Correctness}} 
 & Essential & 719 & 87 & 59 & 8.21\% & 44 & 15 \\
 & Important & 698 & 64 & 49 & 7.02\% & 41 & 8 \\
 & Highlight & 735 & 59 & 42 & 5.71\% & 36 & 6 \\
 & Pitfalls  & 428 & 15 & 10 & 2.34\% & 8 & 2 \\
 & \textbf{Subtotal} & \textbf{2580} & \textbf{225} & \textbf{160} & \textbf{6.20\%} & \textbf{129} & \textbf{31} \\
\midrule
\multirow{5}{*}{\textbf{Comprehensiveness}} 
 & Essential & 719 & 62 & 59 & 8.21\% & 52 & 7 \\
 & Important & 698 & 67 & 54 & 7.74\% & 48 & 6 \\
 & Highlight & 735 & 51 & 45 & 6.12\% & 42 & 3 \\
 & Pitfalls  & 428 & 5  & 3  & 0.70\% & 2  & 1 \\
 & \textbf{Subtotal} & \textbf{2580} & \textbf{185} & \textbf{161} & \textbf{6.24\%} & \textbf{144} & \textbf{17} \\
\midrule
\multirow{5}{*}{\textbf{Relevance}} 
 & Essential & 719 & 2  & 0  & 0.00\% & 0  & 0 \\
 & Important & 698 & 2  & 2  & 0.29\% & 2  & 0 \\
 & Highlight & 735 & 12 & 10 & 1.36\% & 10 & 0 \\
 & Pitfalls  & 428 & 2  & 1  & 0.23\% & 1  & 0 \\
 & \textbf{Subtotal} & \textbf{2580} & \textbf{18}  & \textbf{13}  & \textbf{0.50\%} & \textbf{13}  & \textbf{0} \\
\bottomrule
\end{tabular}%
}
\vspace{1ex}
\parbox{\textwidth}{\footnotesize \textit{Note:} The evaluation baseline consists of 2,580 rubrics derived from 259 queries. ``Preliminary Issues'' refers to anomalies flagged during the independent review phase, whereas ``Confirmed Issues'' denotes valid errors finalized post-consensus adjudication.}
\end{table}

\begin{table}[htbp]
\centering
\caption{Clinical Expert Evaluation of Rubric Weight Rationality}
\label{tab:expert_weight_rationality}
\resizebox{\textwidth}{!}{%
\begin{tabular}{lcccccccc}
\toprule
\multirow{2}{*}{\textbf{Rubric Level}} & \textbf{Total} & \textbf{Preliminary} & \textbf{Confirmed} & \textbf{Confirmed} & \multicolumn{3}{c}{\textbf{Discrepancy Type}} \\
\cmidrule(lr){6-8}
 & \textbf{Rubrics (A)} & \textbf{Discrepancies (B)} & \textbf{Discrepancies (C)} & \textbf{Rate (C/A)} & \textbf{Overestimated} & \textbf{Underestimated} & \textbf{Redundant} \\
\midrule
Essential & 719  & 58  & 13 & 1.81\% & 13 & 0 & 0 \\
Important & 699  & 47  & 16 & 2.29\% & 12 & 3 & 1 \\
Highlight & 734  & 32  & 11 & 1.50\% & 0  & 3 & 8 \\
Pitfalls  & 428  & 8   & 2  & 0.47\% & 0  & 0 & 2 \\
\midrule
\textbf{Total} & \textbf{2580} & \textbf{145} & \textbf{42} & \textbf{1.63\%} & \textbf{25} & \textbf{6} & \textbf{11} \\
\bottomrule
\end{tabular}%
}
\vspace{1ex}
\parbox{\textwidth}{\footnotesize \textit{Note:} Discrepancy types indicate proposed modifications to the system's original ratings. ``Overestimated'' implies marginal information was incorrectly assigned core status; ``Underestimated'' implies core information was assigned marginal status; ``Redundant'' denotes invalid criteria recommended for deletion.}
\end{table}

\subsection{Comprehensive Performance Evaluation of Mainstream Models}
Based on the validated automated evaluation framework detailed previously, this section systematically presents the overall performance and granular capability distribution of various models on QuarkMedBench. This experimental phase aims to establish capability baselines for current large medical models and to deeply analyze the behavioral characteristics and shortcomings of different technical paradigms (e.g., cutting-edge closed-source models, open-weight models, and Chain-of-Thought reasoning models) within real-world medical question-answering scenarios.

\subsubsection{Overall Performance}
To quantitatively and comprehensively evaluate the holistic competence of large models in real-world medical and health question-answering contexts, this study employed stratified random sampling to extract 2,000 core query samples from the entire evaluation dataset for independent testing. The evaluated subjects encompassed representative models across different technical architectures and parameter scales. These included frontier closed-source series (GPT-5 v5.0--v5.2, Gemini 3.0-Pro/2.5-Pro), mainstream open-source series (DeepSeek-V3.2/R1-0528, Kimi-K2, Doubao-1.6), and the Qwen3 series (8B/32B/235B and the CoT-equipped 30B-A3B-Thinking version), which were specifically incorporated to investigate scaling laws.

For metric construction, this study adopted a rubric-based hierarchical quantitative evaluation approach, reflecting capability distribution by calculating average score rates across different standard tiers. The overall score rate, serving as the primary metric for comprehensive performance, is the weighted sum of sub-scores (default weight distribution: Essential:Important:Highlight:Pitfall = 2:1:1:2). Specifically, the \textbf{Essential Standard} primarily measures the model's coverage of core medical information, reflecting its foundational problem-solving capacity. The \textbf{Important Standard} focuses on assessing logical completeness and explanatory depth. The \textbf{Highlight Standard} captures the richness of beneficial supplementary information or extended recommendations provided beyond core facts. Conversely, the \textbf{Pitfall Standard} serves as a quantitative penalty for factual deviations or safety risks within responses; this value is typically negative, with a smaller absolute value denoting higher model safety and reliability.

\begin{table}[htbp]
  \centering
  \caption{Comprehensive Performance of Mainstream LLMs without Output Length Constraints}
  \label{tab:unconstrained_performance}
  \resizebox{\linewidth}{!}{%
  \begin{tabular}{llccccc}
    \toprule
    \multirow{2}{*}{\textbf{Rank}} & \multirow{2}{*}{\textbf{Model}} & \textbf{Avg.} & \textbf{Base} & \multicolumn{3}{c}{\textbf{Ablation Study (Score \%)}} \\
    \cmidrule(lr){5-7}
    & & \textbf{Length} & \textbf{Score (\%)} & \textbf{w/ Saturation} & \textbf{w/ Truncation*} & \textbf{w/ Both*} \\
    \midrule
    1 & GPT-5 & 1,007 & 79.84 & 68.88 & 78.30 & 67.63 \\
    2 & Gemini 2.5 Pro & 1,935 & 74.57 & 66.87 & 72.42 & 65.10 \\
    3 & Qwen3-30B-A3B-Thinking-2507 & 2,044 & 74.54 & 66.15 & 71.64 & 63.72 \\
    4 & DeepSeek-V3.2 & 1,629 & 71.43 & 64.51 & 68.37 & 61.93 \\
    5 & GPT-5.1 & 1,473 & 70.90 & 63.99 & 68.07 & 61.62 \\
    6 & DeepSeek-R1-0528 & 1,388 & 69.96 & 63.97 & 66.83 & 61.34 \\
    7 & Gemini 3.0 Pro & 1,338 & 69.11 & 63.27 & 66.15 & 60.77 \\
    8 & Qwen3-235B-A22B & 1,155 & 67.62 & 61.62 & 63.72 & 58.31 \\
    9 & GPT-5.2 & 788 & 66.19 & 61.09 & 62.69 & 58.11 \\
    10 & Qwen3-32B & 1,269 & 64.91 & 59.35 & 59.77 & 54.93 \\
    11 & Doubao Seed 1.6 & 1,052 & 61.91 & 57.29 & 57.38 & 53.32 \\
    12 & Qwen3-30B-A3B & 1,168 & 58.93 & 55.13 & 53.11 & 49.96 \\
    13 & Qwen3-8B & 1,233 & 57.19 & 53.39 & 50.83 & 47.78 \\
    \bottomrule
  \end{tabular}%
  }
  \vspace{1ex}
  \parbox{\linewidth}{\footnotesize \textit{Note:} Based on 2,000 queries randomly sampled from the full set. Weight distribution is Essential:Important:Highlight:Pitfall = 2:1:1:2. The table is sorted descendingly by the ``w/ Saturation'' column. Columns denoted with * represent ablation study data for the truncation mechanism.}
\end{table}

\begin{table}[htbp]
  \centering
  \caption{Comprehensive Performance of Mainstream LLMs with Strict Length Constraint ($\leq$ 1000 words)}
  \label{tab:constrained_performance}
  \resizebox{\linewidth}{!}{%
  \begin{tabular}{llccccc}
    \toprule
    \multirow{2}{*}{\textbf{Rank}} & \multirow{2}{*}{\textbf{Model}} & \textbf{Avg.} & \textbf{Base} & \multicolumn{3}{c}{\textbf{Ablation Study (Score \%)}} \\
    \cmidrule(lr){5-7}
    & & \textbf{Length} & \textbf{Score (\%)} & \textbf{w/ Saturation} & \textbf{w/ Truncation*} & \textbf{w/ Both*} \\
    \midrule
    1 & GPT-5 & 1,029 & 82.17 & 70.38 & 80.71 & 69.26 \\
    2 & GPT-5.2 & 944 & 72.52 & 65.59 & 67.42 & 61.77 \\
    3 & GPT-5.1 & 1,063 & 69.02 & 63.42 & 65.91 & 60.58 \\
    4 & Qwen3-235B-A22B & 980 & 67.71 & 60.94 & 63.28 & 57.26 \\
    5 & Qwen3-30B-A3B-Thinking-2507 & 977 & 66.23 & 60.10 & 61.59 & 56.14 \\
    6 & Gemini 3.0 Pro & 1,094 & 65.72 & 60.61 & 62.18 & 57.54 \\
    7 & DeepSeek-V3.2 & 1,085 & 64.18 & 59.47 & 58.71 & 54.64 \\
    8 & DeepSeek-R1-0528 & 1,136 & 63.86 & 59.15 & 59.87 & 55.69 \\
    9 & Qwen3-32B & 960 & 63.55 & 57.94 & 58.44 & 53.58 \\
    10 & Gemini 2.5 Pro & 943 & 62.52 & 58.72 & 58.88 & 55.54 \\
    11 & Qwen3-30B-A3B & 1,038 & 60.07 & 55.41 & 54.11 & 50.17 \\
    12 & Doubao Seed 1.6 & 783 & 56.08 & 52.96 & 50.89 & 48.31 \\
    13 & Qwen3-8B & 869 & 53.57 & 50.36 & 46.83 & 44.30 \\
    \bottomrule
  \end{tabular}%
  }
  \vspace{1ex}
  \parbox{\linewidth}{\footnotesize \textit{Note:} Sorted descendingly by Base Score (\%). This table evaluates knowledge distillation under severe information density constraints. *Represents ablation study data for the truncation mechanism.}
\end{table}

\textbf{Overall Performance Analysis:}
\begin{itemize}
    \item \textbf{Evaluation under Unconstrained Conditions:} In the setting without generation length limits (Table~\ref{tab:unconstrained_performance}), the closed-source GPT-5 demonstrated a significant advantage in comprehensive performance, ranking first with a base score of 79.84\%. It was followed by Gemini 2.5 Pro (74.57\%) and the open-weight model equipped with reinforcement learning CoT, Qwen3-30B-A3B-Thinking (74.54\%). The data collectively indicate that the top-tier cohort (base score $>$ 70\%) in the current benchmark is jointly composed of cutting-edge closed-source models and the latest generation of open-source reasoning models (e.g., DeepSeek-V3.2, GPT-5.1).
    \item \textbf{Evaluation under Length-Constrained Scenarios:} Following the imposition of a strict 1,000-word output constraint (Table~\ref{tab:constrained_performance}), the performance ranking among models underwent a substantial transformation. GPT-5.2 exhibited optimal knowledge distillation capability in this constrained environment, surging to the top with a base score of 72.52\%; GPT-5.1 (69.02\%) and Qwen3-235B-A22B (67.71\%) ranked second and third, respectively. Conversely, certain models that excelled in the unconstrained scenario (e.g., GPT-5, Gemini 2.5 Pro) experienced considerable declines in their base scores when subjected to word limits.
    \item \textbf{Calibration Effects of Saturation and Truncation Mechanisms:} Observations from the controlled experiments in both tables reveal that the introduction of the \textbf{saturation mechanism} induced a score callback of 5\% to 10\% for the vast majority of models. This effectively penalized strategies that acquire anomalous scores through tautological repetition or the accumulation of marginal knowledge. Furthermore, the \textbf{truncation mechanism} (which executes circuit-breaker deductions for responses triggering severe medical risks) further widened the performance gap among models. For instance, upon introducing the truncation mechanism in the unconstrained setting, lower-ranked open-source models exhibited significantly sharper score drops compared to leading models, underscoring the vulnerability of foundational models to safety boundary failures during extensive text generation.
\end{itemize}

\subsubsection{Granular Analysis of Evaluation Dimensions}
This section dissects the score distributions across four granular dimensions---Essential, Important, Highlight, and Pitfall (Tables~\ref{tab:granular_unconstrained} and \ref{tab:granular_constrained})---to further elucidate the heterogeneous performance of models regarding knowledge breadth and safety baselines.


\begin{table}[htbp]
  \centering
  \caption{Dimensional Score Distribution under Unconstrained Conditions}
  \label{tab:granular_unconstrained}
  \resizebox{\linewidth}{!}{%
  \begin{tabular}{clcccccc}
    \toprule
    \multirow{2}{*}{\textbf{Rank}} & \multirow{2}{*}{\textbf{Model}} & \textbf{Avg.} & \textbf{Overall} & \multicolumn{4}{c}{\textbf{Dimensional Score Rate (\%)}} \\
    \cmidrule(lr){5-8}
    & & \textbf{Length} & \textbf{Score (\%)} & \textbf{Essential} & \textbf{Important} & \textbf{Highlight} & \textbf{Pitfall} \\
    \midrule
    1 & GPT-5 & 1,007 & 79.84 & 92.23 & 81.54 & 56.04 & -4.30 \\
    2 & Gemini 2.5 Pro & 1,935 & 74.57 & 91.58 & 74.64 & 45.01 & -6.39 \\
    3 & Qwen3-30B-A3B-Thinking-2507 & 2,044 & 74.54 & 89.69 & 74.37 & 49.11 & -6.23 \\
    4 & DeepSeek-V3.2 & 1,629 & 71.43 & 89.48 & 72.29 & 41.86 & -8.66 \\
    5 & GPT-5.1 & 1,473 & 70.90 & 88.40 & 72.32 & 40.37 & -7.78 \\
    6 & DeepSeek-R1-0528 & 1,388 & 69.96 & 88.29 & 70.26 & 38.79 & -7.07 \\
    7 & Gemini 3.0 Pro & 1,338 & 69.11 & 88.36 & 68.64 & 38.75 & -9.32 \\
    8 & Qwen3-235B-A22B & 1,155 & 67.62 & 86.04 & 69.67 & 40.12 & -12.10 \\
    9 & GPT-5.2 & 788 & 66.19 & 84.78 & 65.39 & 33.76 & -5.80 \\
    10 & Qwen3-32B & 1,269 & 64.91 & 83.61 & 68.48 & 38.30 & -14.42 \\
    11 & Doubao Seed 1.6 & 1,052 & 61.91 & 81.64 & 62.31 & 32.30 & -11.20 \\
    12 & Qwen3-30B-A3B & 1,168 & 58.93 & 79.18 & 60.87 & 31.38 & -14.75 \\
    13 & Qwen3-8B & 1,233 & 57.19 & 77.68 & 60.55 & 30.82 & -18.04 \\
    \bottomrule
  \end{tabular}%
  }
\end{table}

\begin{table}[htbp]
  \centering
  \caption{Dimensional Score Distribution with Strict Length Constraint ($\leq$ 1000 words)}
  \label{tab:granular_constrained}
  \resizebox{\linewidth}{!}{%
  \begin{tabular}{clcccccc}
    \toprule
    \multirow{2}{*}{\textbf{Rank}} & \multirow{2}{*}{\textbf{Model}} & \textbf{Avg.} & \textbf{Overall} & \multicolumn{4}{c}{\textbf{Dimensional Score Rate (\%)}} \\
    \cmidrule(lr){5-8}
    & & \textbf{Length} & \textbf{Score (\%)} & \textbf{Essential} & \textbf{Important} & \textbf{Highlight} & \textbf{Pitfall} \\
    \midrule
    1 & GPT-5 & 1,029 & 82.17 & 93.90 & 85.09 & 58.36 & -4.21 \\
    2 & GPT-5.2 & 944 & 72.52 & 89.35 & 73.33 & 41.22 & -5.15 \\
    3 & GPT-5.1 & 1,063 & 69.02 & 87.92 & 69.21 & 36.63 & -7.38 \\
    4 & Qwen3-235B-A22B & 980 & 67.71 & 84.74 & 69.62 & 45.11 & -13.53 \\
    5 & Qwen3-30B-A3B-Thinking-2507 & 977 & 66.23 & 83.49 & 66.57 & 42.93 & -11.81 \\
    6 & Gemini 3.0 Pro & 1,094 & 65.72 & 86.31 & 65.81 & 34.17 & -10.64 \\
    7 & DeepSeek-V3.2 & 1,085 & 64.18 & 84.88 & 63.94 & 32.75 & -10.53 \\
    8 & DeepSeek-R1-0528 & 1,136 & 63.86 & 83.96 & 63.40 & 32.51 & -9.57 \\
    9 & Qwen3-32B & 960 & 63.55 & 83.00 & 67.38 & 39.50 & -18.03 \\
    10 & Gemini 2.5 Pro & 943 & 62.52 & 84.56 & 59.18 & 27.99 & -7.62 \\
    11 & Qwen3-30B-A3B & 1,038 & 60.07 & 79.41 & 63.85 & 34.77 & -16.74 \\
    12 & Doubao Seed 1.6 & 783 & 56.08 & 77.95 & 55.88 & 25.00 & -12.71 \\
    13 & Qwen3-8B & 869 & 53.57 & 74.93 & 57.87 & 28.02 & -20.89 \\
    \bottomrule
  \end{tabular}%
  }
\end{table}

\textbf{Results Analysis:}
\begin{itemize}
    \item \textbf{Convergence in Core Fact Retrieval (Essential Standard):} The data indicate that participating models universally achieved high scores in the Essential dimension, which reflects fundamental medical facts. Under the unconstrained setting, the top seven models all exceeded an 88\% Essential score rate, peaking at 92.23\% (GPT-5). This suggests that current mainstream large language models possess a mature and stable baseline capability regarding the retention and foundational retrieval of static, core medical knowledge.
    \item \textbf{Divergence in Higher-Order Knowledge Extension (Highlight Standard):} In stark contrast to the convergence seen in the Essential standard, models exhibited substantial divergence in the Highlight standard, which captures extended knowledge such as in-depth analysis and customized recommendations. GPT-5's Highlight score rate stabilized within the 56\% to 58\% range, whereas the majority of baseline models hovered between 30\% and 45\%. Notably, this disparity amplified under strict length constraints (widening from 58.36\% to 25.00\%), indicating that executing profound knowledge association within limited text spans remains a technical bottleneck for most current models.
    \item \textbf{Heterogeneity in Medical Errors (Pitfall Standard):} Model performance varied immensely in the Pitfall dimension, which quantifies factual errors and medical risks. Leading models (e.g., GPT-5, GPT-5.2) maintained stringently low penalty rates, confined to a $-$4.2\% to $-$5.8\% range across both experimental settings. A cross-sectional comparison of the Qwen3 series across parameter scales (8B/32B/235B) reveals that, despite marginal differences in their Important standard scores (60\% - 69\%), their Pitfall penalty rates deteriorated precipitously as model size decreased (worsening from $-$9.05\% to $-$18.04\%). This profoundly illustrates the core value of scaling up model parameters in healthcare scenarios: it does not merely augment the knowledge repository but significantly fortifies the model's internal validation and logical interception capabilities against potential medical common-sense errors.
\end{itemize}

\subsubsection{External Benchmark Cross-Validation}
To validate the evaluation efficacy of QuarkMedBench, this study incorporated HealthBench---an authoritative medical benchmark released by OpenAI---as an external reference framework. HealthBench primarily assesses models' medical knowledge retrieval and diagnostic reasoning capabilities based on structured clinical cases. By juxtaposing the performance of participating models across these two evaluation paradigms, we investigated the discriminative power and measurement characteristics of our proposed benchmark.

\begin{table}[htbp]
  \centering
  \caption{Performance Distribution Comparison of Mainstream Models on HealthBench and QuarkMedBench}
  \label{tab:benchmark_comparison}
  \resizebox{\linewidth}{!}{%
  \begin{tabular}{l ccc ccc c}
    \toprule
    \multirow{2}{*}{\textbf{Model}} & \multicolumn{3}{c}{\textbf{HealthBench (HB)}} & \multicolumn{3}{c}{\textbf{QuarkMedBench (QB)}} & \multirow{2}{*}{\textbf{$\Delta$ Rank}} \\
    \cmidrule(lr){2-4} \cmidrule(lr){5-7}
    & \textbf{Score (\%)} & \textbf{Avg. Length} & \textbf{Rank} & \textbf{Score (\%)} & \textbf{Avg. Length} & \textbf{Rank} & \\
    \midrule
    GPT-5            & 65.56 & 1,364 & 1  & 79.84 & 1,007 & 1  & -- \\
    GPT-5.2          & 62.72 & 1,267 & 2  & 66.19 & 788   & 7  & $\downarrow$ 5 \\
    GPT-5.1          & 62.20 & 1,802 & 3  & 70.90 & 1,473 & 4  & $\downarrow$ 1 \\
    Gemini 2.5 Pro   & 55.90 & 1,959 & 4  & 74.57 & 1,935 & 2  & $\uparrow$ 2 \\
    Doubao Seed 1.6  & 54.78 & 1,425 & 5  & 61.91 & 1,052 & 9  & $\downarrow$ 4 \\
    DeepSeek-R1-0528 & 54.74 & 1,473 & 6  & 69.96 & 1,388 & 5  & $\uparrow$ 1 \\
    Gemini 3.0 Pro   & 54.45 & 1,502 & 7  & 69.11 & 1,338 & 6  & $\uparrow$ 1 \\
    Qwen3-32B        & 53.67 & 1,439 & 8  & 64.91 & 1,269 & 8  & -- \\
    DeepSeek-V3.2    & 53.43 & 1,566 & 9  & 71.43 & 1,629 & 3  & $\uparrow$ 6 \\
    Qwen3-8B         & 50.79 & 1,401 & 10 & 57.19 & 1,233 & 10 & -- \\
    \bottomrule
  \end{tabular}%
  }
  \vspace{1ex}
  \parbox{\linewidth}{\footnotesize \textit{Note:} HealthBench metrics preferentially utilize self-tested data reproduced in this study to control experimental variables. QuarkMedBench metrics are derived from the overall score rate under unconstrained conditions. $\Delta$ Rank denotes the ordinal difference between the two benchmarks.}
\end{table}

\textbf{1. Rank Consistency Check} \\
The aggregate data indicate a significant positive correlation in the performance rankings of the evaluated models across both benchmarks. At the extremes of the capability spectrum, the benchmarks demonstrated identical discriminatory power: GPT-5 consistently secured the top position in both evaluations, while the smaller-parameter Qwen3-8B consistently ranked last. Concurrently, the rank fluctuations for the majority of mid-tier models (e.g., GPT-5.1, DeepSeek-R1, Gemini 3.0 Pro) were constrained to a narrow margin of 1 to 2 positions. This highly concordant ranking pattern empirically validates that QuarkMedBench possesses equivalent efficacy and discriminative capacity to authoritative closed-domain benchmarks (i.e., HealthBench) in assessing the foundational medical competence of large language models.

\textbf{2. Score Shift Phenomenon} \\
In terms of absolute metrics, the overall score rates of models on QuarkMedBench (57\%--80\%) systematically superseded those on HealthBench (50\%--66\%). This score shift primarily stems from fundamental differences in task design: HealthBench emphasizes the retrieval of a singular, deterministic answer in an objective-test format, offering minimal margin for error. Conversely, QuarkMedBench employs a multi-dimensional, fine-grained scoring rubric framework. This evaluation schema not only encompasses core medical facts (Essential standard) but also rewards informational completeness and extended recommendations (Important and Highlight standards). Consequently, the elevated absolute scores do not imply reduced task difficulty; rather, they reflect the broader accommodation of comprehensive generative capabilities inherent in open-ended question-answering.

\textbf{3. Specificity Analysis of Rank Fluctuation} \\
Against the backdrop of overall rank consistency, a minority of models exhibited substantial positional inversions (e.g., GPT-5.2 dropping 5 places, DeepSeek-V3.2 rising 6 places). A comparative analysis of text generation lengths reveals that these fluctuations correlate with the models' distinct output strategies under varying instructional environments. When processing the structured inputs of HealthBench, GPT-5.2 gravitated towards exhaustive discourse (averaging 1,267 words); however, when confronted with the unstructured user queries of QuarkMedBench, its output strategy contracted significantly (averaging 788 words), resulting in point deductions due to insufficient coverage of long-tail knowledge. Conversely, DeepSeek-V3.2 maintained a high volume of text generation ($>$1,500 words) across both tasks. This divergence objectively demonstrates that, beyond measuring static medical knowledge, QuarkMedBench effectively captures a model's response strategy and informational exhaustiveness when addressing real-world, long-tail queries, thereby serving as a robust evaluative complement to traditional medical benchmarks.

\subsection{Ablation Study of Key Influencing Factors}
\label{sec:ablation_study}

Based on the aforementioned experimental data, this section employs a control variable method to systematically analyze the specific impacts of text length, reasoning paradigms (Chain of Thought), and parameter scale on the clinical QA performance of LLMs.

\subsubsection{Correlation Between Text Length and Information Density (Length Bias)}
Generative LLMs universally exhibit Length Bias, a phenomenon where models artificially inflate evaluation metrics by generating excessively long texts. To evaluate the true information distillation capabilities of models, we compared their performance under unconstrained conditions versus strict length constraints ($\le 1000$ tokens).

Experimental data reveals significant differences in models' sensitivity to length constraints. Under unconstrained conditions, Gemini 2.5 Pro averages 1,935 tokens per response with an overall score of 74.57\%; when length constraints are applied, its average length drops to 943 tokens, and its score simultaneously decreases to 62.52\%. This indicates that a substantial portion of this model's score in open-ended QA relies on textual expansion, with limited information density per unit of text. In contrast, GPT-5.1 averages 1,473 tokens unconstrained (scoring 70.90\%) and maintains a stable score of 69.02\% when constrained (1,063 tokens). Similarly, Qwen3-235B remains stable (shifting marginally from 67.62\% to 67.71\%) under constraints. These results demonstrate that certain models can effectively adjust their generation strategies under constraints to preserve the integrity of core medical facts. In clinical assistance scenarios, models with high information density convey critical information with lower textual redundancy, yielding higher practical application value.

\subsubsection{Value Validation of the Chain of Thought (CoT) Mechanism}
To investigate the practical role of the Chain of Thought (CoT) mechanism in complex medical decision-making, we compared the performance of the same base models under CoT reasoning versus standard instruction modes.

Taking the Qwen3-30B series as an example, the CoT-enabled Qwen3-30B-Thinking achieved an overall score of 74.54\% under unconstrained conditions, significantly outperforming the baseline model Qwen3-30B (58.93\%). Fine-grained analysis reveals that the performance lift from CoT manifests primarily in two dimensions: 
\textbf{1) Significant Reduction in Error Rates:} The Pitfall deduction rate for the Thinking model is $-6.23\%$, compared to $-14.75\%$ for the baseline. This indicates that the internal logical verification inherent in CoT significantly reduces the probability of factual errors and dangerous advice. 
\textbf{2) Enhancement in High-order Knowledge Association:} In the Highlight dimension, the Thinking model (49.11\%) vastly outperforms the baseline (31.38\%). Although CoT significantly increases the average output length (from 1,168 to 2,044 tokens), its stable performance under the saturation mechanism proves that the increased computational overhead translates into effective clinical reasoning rather than meaningless textual repetition.

\subsubsection{Scaling Effect of Model Parameter Size}
To evaluate the impact of parameter scale on medical knowledge recall and complex instruction adherence, we conducted a horizontal comparison of four versions within the Qwen3 series (8B, 30B, 32B, 235B).

Results indicate a positive correlation between the overall score rate and model size: Qwen3-8B (57.19\%) $<$ Qwen3-30B (58.93\%) $<$ Qwen3-32B (64.91\%) $<$ Qwen3-235B (67.62\%). In terms of fine-grained capabilities, expanding the parameter scale yields significant marginal benefits in controlling the Pitfall standard. The deduction rates for the 235B ($-12.10\%$) and 32B ($-14.42\%$) models are notably lower than that of the 8B model ($-18.04\%$).

\subsubsection{Strategic Adaptability Analysis of Generation Length}
Building upon the quantification of length bias, we further analyzed the models' capability to dynamically adjust generation length based on task constraints. Real-world medical applications require models to transcend fixed generation tendencies and exhibit adaptability in information density. For instance, quick consultation scenarios demand core conclusions with low cognitive load, whereas complex case analyses rely on exhaustive medical mechanism explanations to build patient trust. 

Comparing the score structures between unconstrained and strict constraint ($\le 1000$ tokens) conditions reveals that some models can flexibly allocate information distribution. For example, under strict constraints, GPT-5.2 prioritizes a high Essential score (89.35\%) while rationally compressing the length of Important and Highlight items. This indicates the model possesses the strategic adaptability to allocate limited textual space primarily to core medical facts, preventing the loss of critical information due to physical truncation.

\subsection{Rubrics as Rewards: Guiding RL Alignment}
\label{sec:rubrics_as_rewards}

Beyond serving as a standardized static evaluation tool, this study further explores the effectiveness of automated scoring rubrics (Auto-Rubrics) as fine-grained reward signals for reinforcement learning (RL) alignment. In traditional RL alignment paradigms, reward signals are frequently overly sparse or generalized, failing to adequately capture the nuanced coverage of specific medical knowledge points. By deconstructing the Rubric evaluations, we establish an objective, highly granular feedback mechanism for model optimization.

\subsubsection{Construction of Reward Signals}
Although Rubrics excel at measuring medical point coverage, they exhibit inherent limitations in capturing absolute factual correctness, logical coherence, and formatting standardization. Consequently, we adopt the GRPO (Group Relative Policy Optimization) algorithm to construct a hybrid reward function. 

\textbf{Multi-dimensional Signal Fusion:} The final reward signal is formulated as a linear weighted combination of the Rubric Score (focusing on medical point coverage), Helpfulness Reward (focusing on instruction adherence and readability), and Honesty Reward (focusing on factual accuracy). \\
\textbf{Complementary Mechanism:} The Helpfulness and Honesty Reward Models (RMs) effectively compensate for the evaluation blind spots of Rubrics in unstructured dimensions. This synergy ensures that the model can internalize fine-grained scoring feedback without triggering distribution collapse caused by reward hacking, thereby constituting a highly robust training guidance signal.

\subsubsection{Activation and Scaling Mechanisms}

Building upon the foundational scoring logic, we introduce activation and dynamic scaling mechanisms to optimize RL training trajectories. The rationale behind \textbf{activation} is conceptually analogous to the ReLU activation function in neural networks: during pairwise scoring evaluation, only score differentials exceeding a predefined threshold ($L$) are considered valid (Equation~\ref{eq:activation}). This mechanism is particularly advantageous in RL, as activation control enables the selection of rollout samples with a substantial margin, thereby preventing the policy from converging to local optima.

Furthermore, when utilizing Rubric scores to guide training, the absolute magnitude of score improvement is intrinsically linked to the current baseline score. As the baseline score approaches saturation, the model should be incentivized to explore marginal improvements, which must be disproportionately rewarded (e.g., the difficulty of improving from 0.8 to 0.9 points significantly exceeds that of moving from 0.2 to 0.3 points). To address this, we implement a \textbf{scaling} methodology (Equation~\ref{eq:scaling}): if a sampling result exceeds the collective mean ($S_{mean}$), its reward score is amplified to maximize the Reward Margin. 

\begin{equation}
Equal_{(A,B)} =
\begin{cases}
\text{True}, & |S_A - S_B| < L \\
\text{False}, & \text{otherwise}
\end{cases}
\label{eq:activation}
\end{equation}

\begin{equation}
S_{mean} = \frac{\sum_{i \in \{GPT, R1, Qwen, \dots\}} S_i}{N}
\label{eq:mean}
\end{equation}

\begin{equation}
S_{scale} =
\begin{cases}
S_{roll}, & S_{roll} < S_{mean} \\
S_{mean} + W_{scale}(S_{roll} - S_{mean}), & S_{roll} \ge S_{mean}
\end{cases}
\label{eq:scaling}
\end{equation}

Ablation studies corroborate the necessity of these specialized mechanisms. Experimental data reveal that upon removing the activation threshold and dynamic scaling parameters, the model's ranking accuracy metric (correctly ranked/reversed pair ratio) in evaluation tasks deteriorates precipitously from 3.5 down to 2.3. This underscores that filtering low-value stochastic fluctuations and amplifying the reward margin for high-quality samples are critical imperatives for guiding the model toward higher-order clinical reasoning capabilities.

\textbf{For an exhaustive exposition on the algorithmic specifics and the concrete empirical effects of utilizing "Rubrics-As-Reward" to guide medical model training, please refer to the specialized technical report on alignment algorithms authored by the Quark Medical Team \cite{xu_quark_2026}.}

\section{Discussion and Limitations}
\label{sec:discussion}

By constructing QuarkMedBench, this study validates the feasibility of using multi-model consensus-based automated evaluation to replace human assessment in the vertical medical domain. However, as an evaluation system simulating real clinical interactions, several limitations remain regarding dimensional completeness, dynamic fidelity, and endogenous algorithmic biases.

\textbf{Limitations in Empathy and Understandability:} Current scoring rubrics rely heavily on extracting knowledge points from gold standards, which incentivizes models to stack high-density professional terminology to maximize hit rates. As noted by Gu et al. (2025), high medical accuracy does not equate to high user understandability \cite{yao_biased_2025}. Models may generate jargon-heavy responses, neglecting the general public's need for layperson-friendly explanations and empathic emotional support—elements that are difficult to quantify via discrete knowledge points.

\textbf{Lack of Dynamic Interactive Simulation:} The current framework deconstructs multi-turn dialogues into independent single-turn queries. While this accurately measures response quality at a specific node, it fails to quantify a model's contextual tracking ability as user intents evolve \cite{manczak_shallow_2025}. Real-world medical consultations often require proactive inquiry to acquire missing information. Our passive-response mechanism lacks dynamic Agent-based game theory simulations \cite{almansoori_self-evolving_2025}, which is an urgent dimension to supplement in future medical evaluations.

\textbf{Endogenous Length Bias and Retrieval Dependency:} Although our hard truncation mechanism (e.g., 1000 tokens) mitigates artificial score inflation caused by verbose redundancy, it acts as a heuristic post-processing correction rather than fundamentally altering the model's underlying preference for long texts. Future work should explore Semantic Prompting Strategies \cite{hu_explaining_2025} to intrinsically incentivize high-density expression. Additionally, the reliability of our Auto-Rubrics relies heavily on the quality of DeepResearch. If the external retrieval pool harbors systematic biases or lacks authoritative Chinese evidence for rare diseases, factual deviations may still occur. 

\section{Conclusion}
\label{sec:conclusion}

This study introduces \textit{QuarkMedBench}, an LLM evaluation benchmark designed to bridge the gap between static medical knowledge assessments and real-world clinical applications. By reconstructing unstructured, ambiguous, and long-tail queries from actual online healthcare scenarios, we established a complex task system covering clinical care, wellness health, and professional inquiries. 

Methodologically, we proposed a fine-grained automated evaluation framework based on multi-model consensus, decoupling open-ended responses into Essential, Important, Highlight, and Pitfall dimensions. By integrating weight-based truncation and saturation mechanisms, we effectively calibrated scoring biases caused by textual redundancy. Empirical analysis across mainstream LLMs highlights their capability boundaries, exposing prevalent length biases while validating the effectiveness of Chain of Thought (CoT) reasoning and model scaling in suppressing medical hallucinations. 

Crucially, the high-granularity Auto-Rubrics generated in this study transcend static evaluation, proving their viability as fine-grained reward signals for Reinforcement Learning (Rubrics as Rewards). Moving forward, medical AI evaluation must evolve from static paradigms towards multi-modal dynamic interactions and Agent-based clinical simulations. QuarkMedBench establishes a standardized baseline for medical AI safety and utility, illuminating an empirical trajectory for the next generation of algorithmic iteration and evaluation systems.

\bibliography{references}

\begin{thebibliography}{21}
\providecommand{\natexlab}[1]{#1}
\providecommand{\url}[1]{\texttt{#1}}
\expandafter\ifx\csname urlstyle\endcsname\relax
  \providecommand{\doi}[1]{doi: #1}\else
  \providecommand{\doi}{doi: \begingroup \urlstyle{rm}\Url}\fi

\bibitem[Nori et~al.(2023)Nori, King, McKinney, Carignan, and Horvitz]{nori_capabilities_2023}
Harsha Nori, Nicholas King, Scott~Mayer McKinney, Dean Carignan, and Eric Horvitz.
\newblock Capabilities of {GPT}-4 on {Medical} {Challenge} {Problems}, April 2023.
\newblock URL \url{http://arxiv.org/abs/2303.13375}.
\newblock arXiv:2303.13375 [cs].

\bibitem[Singhal et~al.(2023)Singhal, Azizi, Tu, Mahdavi, Wei, Chung, Scales, Tanwani, Cole-Lewis, Pfohl, Payne, Seneviratne, Gamble, Kelly, Babiker, Schärli, Chowdhery, Mansfield, Demner-Fushman, Agüera Y~Arcas, Webster, Corrado, Matias, Chou, Gottweis, Tomasev, Liu, Rajkomar, Barral, Semturs, Karthikesalingam, and Natarajan]{singhal_large_2023}
Karan Singhal, Shekoofeh Azizi, Tao Tu, S.~Sara Mahdavi, Jason Wei, Hyung~Won Chung, Nathan Scales, Ajay Tanwani, Heather Cole-Lewis, Stephen Pfohl, Perry Payne, Martin Seneviratne, Paul Gamble, Chris Kelly, Abubakr Babiker, Nathanael Schärli, Aakanksha Chowdhery, Philip Mansfield, Dina Demner-Fushman, Blaise Agüera Y~Arcas, Dale Webster, Greg~S. Corrado, Yossi Matias, Katherine Chou, Juraj Gottweis, Nenad Tomasev, Yun Liu, Alvin Rajkomar, Joelle Barral, Christopher Semturs, Alan Karthikesalingam, and Vivek Natarajan.
\newblock Large language models encode clinical knowledge.
\newblock \emph{Nature}, 620\penalty0 (7972):\penalty0 172--180, August 2023.
\newblock ISSN 1476-4687.
\newblock \doi{10.1038/s41586-023-06291-2}.

\bibitem[Singhal et~al.(2025)Singhal, Tu, Gottweis, Sayres, Wulczyn, Amin, Hou, Clark, Pfohl, Cole-Lewis, Neal, Rashid, Schaekermann, Wang, Dash, Chen, Shah, Lachgar, Mansfield, Prakash, Green, Dominowska, Agüera Y~Arcas, Tomašev, Liu, Wong, Semturs, Mahdavi, Barral, Webster, Corrado, Matias, Azizi, Karthikesalingam, and Natarajan]{singhal_toward_2025}
Karan Singhal, Tao Tu, Juraj Gottweis, Rory Sayres, Ellery Wulczyn, Mohamed Amin, Le~Hou, Kevin Clark, Stephen~R. Pfohl, Heather Cole-Lewis, Darlene Neal, Qazi~Mamunur Rashid, Mike Schaekermann, Amy Wang, Dev Dash, Jonathan~H. Chen, Nigam~H. Shah, Sami Lachgar, Philip~Andrew Mansfield, Sushant Prakash, Bradley Green, Ewa Dominowska, Blaise Agüera Y~Arcas, Nenad Tomašev, Yun Liu, Renee Wong, Christopher Semturs, S.~Sara Mahdavi, Joelle~K. Barral, Dale~R. Webster, Greg~S. Corrado, Yossi Matias, Shekoofeh Azizi, Alan Karthikesalingam, and Vivek Natarajan.
\newblock Toward expert-level medical question answering with large language models.
\newblock \emph{Nature Medicine}, 31\penalty0 (3):\penalty0 943--950, March 2025.
\newblock ISSN 1546-170X.
\newblock \doi{10.1038/s41591-024-03423-7}.

\bibitem[Liu et~al.(2024)Liu, Li, Zhou, Yin, Yang, Tang, Luo, Zeng, Jiang, Gao, Nigam, Nag, Yin, Hua, Zhou, Rohanian, Thakur, Clifton, and Clifton]{liu_large_2024}
Fenglin Liu, Zheng Li, Hongjian Zhou, Qingyu Yin, Jingfeng Yang, Xianfeng Tang, Chen Luo, Ming Zeng, Haoming Jiang, Yifan Gao, Priyanka Nigam, Sreyashi Nag, Bing Yin, Yining Hua, Xuan Zhou, Omid Rohanian, Anshul Thakur, Lei Clifton, and David~A. Clifton.
\newblock Large {Language} {Models} {Are} {Poor} {Clinical} {Decision}-{Makers}: {A} {Comprehensive} {Benchmark}.
\newblock In Yaser Al-Onaizan, Mohit Bansal, and Yun-Nung Chen, editors, \emph{Proceedings of the 2024 {Conference} on {Empirical} {Methods} in {Natural} {Language} {Processing}}, pages 13696--13710, Miami, Florida, USA, November 2024. Association for Computational Linguistics.
\newblock \doi{10.18653/v1/2024.emnlp-main.759}.
\newblock URL \url{https://aclanthology.org/2024.emnlp-main.759/}.

\bibitem[He et~al.(2025)He, Mao, Lin, Ruan, Lan, Feng, and Cambria]{he_survey_2025}
Kai He, Rui Mao, Qika Lin, Yucheng Ruan, Xiang Lan, Mengling Feng, and Erik Cambria.
\newblock A {Survey} of {Large} {Language} {Models} for {Healthcare}: from {Data}, {Technology}, and {Applications} to {Accountability} and {Ethics}, January 2025.
\newblock URL \url{http://arxiv.org/abs/2310.05694}.
\newblock arXiv:2310.05694 [cs].

\bibitem[Jin et~al.(2021)Jin, Pan, Oufattole, Weng, Fang, and Szolovits]{jin_what_2021}
Di~Jin, Eileen Pan, Nassim Oufattole, Wei-Hung Weng, Hanyi Fang, and Peter Szolovits.
\newblock What {Disease} {Does} {This} {Patient} {Have}? {A} {Large}-{Scale} {Open} {Domain} {Question} {Answering} {Dataset} from {Medical} {Exams}.
\newblock \emph{Applied Sciences}, 11\penalty0 (14), July 2021.
\newblock ISSN 2076-3417.
\newblock \doi{10.3390/app11146421}.
\newblock URL \url{https://www.mdpi.com/2076-3417/11/14/6421}.

\bibitem[Pal et~al.(2022)Pal, Umapathi, and Sankarasubbu]{pal_medmcqa_2022}
Ankit Pal, Logesh~Kumar Umapathi, and Malaikannan Sankarasubbu.
\newblock {MedMCQA} : {A} {Large}-scale {Multi}-{Subject} {Multi}-{Choice} {Dataset} for {Medical} domain {Question} {Answering}, March 2022.
\newblock URL \url{http://arxiv.org/abs/2203.14371}.
\newblock arXiv:2203.14371 [cs].

\bibitem[Jin et~al.(2019)Jin, Dhingra, Liu, Cohen, and Lu]{jin_pubmedqa_2019}
Qiao Jin, Bhuwan Dhingra, Zhengping Liu, William Cohen, and Xinghua Lu.
\newblock {PubMedQA}: {A} {Dataset} for {Biomedical} {Research} {Question} {Answering}.
\newblock In Kentaro Inui, Jing Jiang, Vincent Ng, and Xiaojun Wan, editors, \emph{Proceedings of the 2019 {Conference} on {Empirical} {Methods} in {Natural} {Language} {Processing} and the 9th {International} {Joint} {Conference} on {Natural} {Language} {Processing} ({EMNLP}-{IJCNLP})}, pages 2567--2577, Hong Kong, China, November 2019. Association for Computational Linguistics.
\newblock \doi{10.18653/v1/D19-1259}.
\newblock URL \url{https://aclanthology.org/D19-1259/}.

\bibitem[Wu et~al.(2025)Wu, Wang, and Qin]{wu_performance_2025}
Jin Wu, Zhiheng Wang, and Yifan Qin.
\newblock Performance of {DeepSeek}-{R1} and {ChatGPT}-4o on the {Chinese} {National} {Medical} {Licensing} {Examination}: {A} {Comparative} {Study}.
\newblock \emph{Journal of Medical Systems}, 49\penalty0 (1):\penalty0 74, June 2025.
\newblock ISSN 1573-689X.
\newblock \doi{10.1007/s10916-025-02213-z}.

\bibitem[Li et~al.(2025)Li, Yan, Cai, Li, Zhao, Yao, Liu, Jiang, Xu, Dong, Sun, Zhang, Gui, Liu, Shang, Wu, Cao, Ma, and Jia]{li_quarkmed_2025}
Ao~Li, Bin Yan, Bingfeng Cai, Chenxi Li, Cunzhong Zhao, Fugen Yao, Gaoqiang Liu, Guanjun Jiang, Jian Xu, Liang Dong, Liansheng Sun, Rongshen Zhang, Xiaolei Gui, Xin Liu, Xin Shang, Yao Wu, Yu~Cao, Zhenxin Ma, and Zhuang Jia.
\newblock {QuarkMed} {Medical} {Foundation} {Model} {Technical} {Report}, August 2025.
\newblock URL \url{http://arxiv.org/abs/2508.11894}.
\newblock arXiv:2508.11894 [cs].

\bibitem[Thirunavukarasu et~al.(2023)Thirunavukarasu, Ting, Elangovan, Gutierrez, Tan, and Ting]{thirunavukarasu_large_2023}
Arun~James Thirunavukarasu, Darren Shu~Jeng Ting, Kabilan Elangovan, Laura Gutierrez, Ting~Fang Tan, and Daniel Shu~Wei Ting.
\newblock Large language models in medicine.
\newblock \emph{Nature Medicine}, 29\penalty0 (8):\penalty0 1930--1940, August 2023.
\newblock ISSN 1546-170X.
\newblock \doi{10.1038/s41591-023-02448-8}.

\bibitem[Arora et~al.(2025)Arora, Wei, Hicks, Bowman, Quiñonero-Candela, Tsimpourlas, Sharman, Shah, Vallone, Beutel, Heidecke, and Singhal]{arora_healthbench_2025}
Rahul~K. Arora, Jason Wei, Rebecca~Soskin Hicks, Preston Bowman, Joaquin Quiñonero-Candela, Foivos Tsimpourlas, Michael Sharman, Meghan Shah, Andrea Vallone, Alex Beutel, Johannes Heidecke, and Karan Singhal.
\newblock {HealthBench}: {Evaluating} {Large} {Language} {Models} {Towards} {Improved} {Human} {Health}, May 2025.
\newblock URL \url{http://arxiv.org/abs/2505.08775}.
\newblock arXiv:2505.08775 [cs].

\bibitem[Gunjal et~al.(2025)Gunjal, Wang, Lau, Nath, He, Liu, and Hendryx]{gunjal_rubrics_2025}
Anisha Gunjal, Anthony Wang, Elaine Lau, Vaskar Nath, Yunzhong He, Bing Liu, and Sean Hendryx.
\newblock Rubrics as {Rewards}: {Reinforcement} {Learning} {Beyond} {Verifiable} {Domains}, October 2025.
\newblock URL \url{http://arxiv.org/abs/2507.17746}.
\newblock arXiv:2507.17746 [cs].

\bibitem[Huang et~al.(2025)Huang, Zhuang, Lu, Qin, Xu, Zhao, Peng, Hu, Shen, Hu, Gu, Tu, Liu, Chen, Fu, Fan, Gu, Wang, Yang, Li, and Zhao]{huang_reinforcement_2025}
Zenan Huang, Yihong Zhuang, Guoshan Lu, Zeyu Qin, Haokai Xu, Tianyu Zhao, Ru~Peng, Jiaqi Hu, Zhanming Shen, Xiaomeng Hu, Xijun Gu, Peiyi Tu, Jiaxin Liu, Wenyu Chen, Yuzhuo Fu, Zhiting Fan, Yanmei Gu, Yuanyuan Wang, Zhengkai Yang, Jianguo Li, and Junbo Zhao.
\newblock Reinforcement {Learning} with {Rubric} {Anchors}, August 2025.
\newblock URL \url{http://arxiv.org/abs/2508.12790}.
\newblock arXiv:2508.12790 [cs].

\bibitem[Croxford et~al.(2025)Croxford, Gao, First, Pellegrino, Schnier, Caskey, Oguss, Wills, Chen, Dligach, Churpek, Mayampurath, Liao, Goswami, Wong, Patterson, and Afshar]{croxford_automating_2025}
Emma Croxford, Yanjun Gao, Elliot First, Nicholas Pellegrino, Miranda Schnier, John Caskey, Madeline Oguss, Graham Wills, Guanhua Chen, Dmitriy Dligach, Matthew~M Churpek, Anoop Mayampurath, Frank Liao, Cherodeep Goswami, Karen~K. Wong, Brian~W. Patterson, and Majid Afshar.
\newblock Automating {Evaluation} of {AI} {Text} {Generation} in {Healthcare} with a {Large} {Language} {Model} ({LLM})-as-a-{Judge}.
\newblock \emph{medRxiv}, page 2025.04.22.25326219, May 2025.
\newblock \doi{10.1101/2025.04.22.25326219}.
\newblock URL \url{https://pmc.ncbi.nlm.nih.gov/articles/PMC12045442/}.

\bibitem[Zhang et~al.(2025)Zhang, Li, Cui, Cai, Liu, Fu, Huang, Zhao, Zhang, Chen, Wang, Luu, Bi, Shi, and Shi]{zhang_sirens_2025}
Yue Zhang, Yafu Li, Leyang Cui, Deng Cai, Lemao Liu, Tingchen Fu, Xinting Huang, Enbo Zhao, Yu~Zhang, Yulong Chen, Longyue Wang, Anh~Tuan Luu, Wei Bi, Freda Shi, and Shuming Shi.
\newblock Siren’s {Song} in the {AI} {Ocean}: {A} {Survey} on {Hallucination} in {Large} {Language} {Models}.
\newblock \emph{Computational Linguistics}, 51\penalty0 (4):\penalty0 1373--1418, December 2025.
\newblock ISSN 0891-2017.
\newblock \doi{10.1162/COLI.a.16}.
\newblock URL \url{https://doi.org/10.1162/COLI.a.16}.

\bibitem[Xu et~al.(2026)Xu, Liu, Tong, Xu, Wei, Feng, Hou, Yin, Hu, Zhou, Ma, Xu, and Jiang]{xu_quark_2026}
Tianxiang Xu, Jiayi Liu, Yixuan Tong, Jialu Xu, Yunqing Wei, Kaiwen Feng, PanPan Hou, Kangping Yin, Jiyuan Hu, Hao Zhou, Zhenxin Ma, Jian Xu, and Guanjun Jiang.
\newblock Quark {Medical} {Alignment}: {A} {Holistic} {Multi}-{Dimensional} {Alignment} and {Collaborative} {Optimization} {Paradigm}, March 2026.
\newblock URL \url{http://arxiv.org/abs/2602.11661}.
\newblock arXiv:2602.11661 [cs].

\bibitem[Yao et~al.(2025)Yao, Liu, Drui, Pettersson, Blasimme, and Kijewski]{yao_biased_2025}
Jianzhou Yao, Shunchang Liu, Guillaume Drui, Rikard Pettersson, Alessandro Blasimme, and Sara Kijewski.
\newblock The {Biased} {Oracle}: {Assessing} {LLMs}' {Understandability} and {Empathy} in {Medical} {Diagnoses}, November 2025.
\newblock URL \url{http://arxiv.org/abs/2511.00924}.
\newblock arXiv:2511.00924 [cs].

\bibitem[Manczak et~al.(2025)Manczak, Lin, Eiras, Neill, and Mugunthan]{manczak_shallow_2025}
Blazej Manczak, Eric Lin, Francisco Eiras, James~O' Neill, and Vaikkunth Mugunthan.
\newblock Shallow {Robustness}, {Deep} {Vulnerabilities}: {Multi}-{Turn} {Evaluation} of {Medical} {LLMs}, October 2025.
\newblock URL \url{http://arxiv.org/abs/2510.12255}.
\newblock arXiv:2510.12255 [cs].

\bibitem[Almansoori et~al.(2025)Almansoori, Kumar, and Cholakkal]{almansoori_self-evolving_2025}
Mohammad Almansoori, Komal Kumar, and Hisham Cholakkal.
\newblock Self-{Evolving} {Multi}-{Agent} {Simulations} for {Realistic} {Clinical} {Interactions}, October 2025.
\newblock URL \url{http://arxiv.org/abs/2503.22678}.
\newblock arXiv:2503.22678 [cs].

\bibitem[Hu et~al.(2025)Hu, Song, Zhang, Xiao, Wang, Chen, Yuan, Lian, Ding, and Xiong]{hu_explaining_2025}
Zhengyu Hu, Linxin Song, Jieyu Zhang, Zheyuan Xiao, Tianfu Wang, Zhengyu Chen, Nicholas~Jing Yuan, Jianxun Lian, Kaize Ding, and Hui Xiong.
\newblock Explaining {Length} {Bias} in {LLM}-{Based} {Preference} {Evaluations}.
\newblock In Christos Christodoulopoulos, Tanmoy Chakraborty, Carolyn Rose, and Violet Peng, editors, \emph{Findings of the {Association} for {Computational} {Linguistics}: {EMNLP} 2025}, pages 6763--6794, Suzhou, China, November 2025. Association for Computational Linguistics.
\newblock ISBN 9798891763357.
\newblock \doi{10.18653/v1/2025.findings-emnlp.358}.
\newblock URL \url{https://aclanthology.org/2025.findings-emnlp.358/}.

\end{thebibliography}

\section*{Appendix}
\appendix

\section{Examples of Ground Truth and Rubrics}

This section elucidates the structural composition and clinical depth of the Ground Truth (GT) and the corresponding automated scoring rubrics generated by our pipeline. By presenting tangible examples across diverse medical scenarios, we demonstrate the efficacy of the multi-model consensus mechanism in distilling complex medical knowledge into deterministic evaluation criteria.

\subsection{Example Ground Truth}

The following JSON snippet presents a representative example of the synthesized Ground Truth for a query regarding hypertension diagnostic standards. It demonstrates how responses from multiple advanced LLMs are aggregated and deconstructed to form a comprehensive, multi-dimensional medical consensus. The information is meticulously categorized into essential facts (e.g., core diagnostic criteria), clinical highlights (e.g., variations in international guidelines), and supplementary extensions (e.g., lifestyle interventions), serving as the foundational corpus for subsequent rubric generation.

\begin{figure}[htbp]
\centering
\begin{jsonbox}[title={\footnotesize\textbf{Code Snippet 1: Multidimensional Ground Truth Schema (Hypertension Example)}}, 
                fonttitle=\sffamily\bfseries, coltitle=black, colbacktitle=jsonbg!90!white]
{
  "query": "What are the diagnostic criteria for hypertension?",
  "merge_spo_info": {
    "Qwen3-Max": {
      "imp": [
        "1. Diagnostic Criteria: The 2023 Chinese Guidelines and US AHA/ACC define clinic BP >=130/80 mmHg as hypertension.",
        "2. Guideline Variations: European ESC/ESH and WHO guidelines maintain >=140/90 mmHg as the threshold for diagnosis.",
        "3. Measurement Protocols: Requires resting for 5 minutes; an average of 2-3 readings taken on non-consecutive days."
      ],
      "aha": [
        "1. Clinical Trends & Controversies: Lowering the threshold to 130/80 mmHg emphasizes early intervention. Conversely, retaining the 140/90 mmHg standard reflects concerns over over-diagnosis.",
        "2. Diagnostic Gold Standards: Ambulatory Blood Pressure Monitoring (ABPM) is essential to rule out 'White-coat' hypertension."
      ],
      "ext": [
        "1. Pre-measurement Interference: Avoid intense exercise, caffeine, and smoking within 30 minutes prior to measurement.",
        "2. Risk Stratification: Routinely assess target organ damage (e.g., left ventricular hypertrophy) and exclude secondary hypertension."
      ]
    },
    "GPT-5": {
      "imp": [
        "1. Diagnostic Criteria (Clinic BP): >=130/80 mmHg (China 2023 and US AHA), requiring multiple non-same-day measurements.",
        "[... Redundant consensus items successfully merged & folded ...]"
      ],
      "aha": [
        "[... Cross-validated insights folded ...]"
      ]
    }
  }
}
\end{jsonbox}
\vspace{-2mm}
\caption{An empirical example of the synthesized Ground Truth. The JSON schema demonstrates the pipeline's capability to decompose complex medical concepts into three granular dimensions: essential facts (\texttt{imp}), deep clinical insights (\texttt{aha}), and extended knowledge (\texttt{ext}), while successfully capturing discrepancies across international clinical guidelines.}
\label{fig:gt_schema}
\end{figure}

\subsection{Example of Rubrics-1}
The ensuing examples illustrate the fine-grained, structured evaluation rubrics automatically transfigured from the synthesized Ground Truth. Each rubric delineates strict, multi-tiered grading Rubric across Essential, Important, Highlight, and Pitful dimensions. By transforming qualitative medical narratives into deterministic, quantitative rules, these rubrics ensure transparent, reproducible, and rigorous capability assessments across varied clinical scenarios (e.g., pharmacological inquiries, dermatological triage, and oncological screening).

\begin{lstlisting}[language=json, style=promptstyle]
{
  "Evaluation_System": {
    "Query": "Eucalyptol, Limonene, and Pinene Enteric Soft Capsules",
    "Department": "Respiratory Medicine",
    "Essential": [
      {
        "Title": "Ingredient & Formulation Assessment",
        "Weight": 2,
        "Description": {
          "Not Met": "Fails to mention the primary active ingredients or the enteric soft capsule formulation.",
          "Partially Met": "Mentions primary ingredients or the enteric formulation, but omits the explanation that the enteric coating is designed to bypass gastric irritation and release in the small intestine.",
          "Fully Met": "Accurately identifies the medication as a compound preparation. Explicitly elucidates that the enteric soft capsule is designed to bypass gastric acid, releasing in the alkaline environment of the small intestine to minimize gastric irritation and optimize absorption."
        }
      },
      {
        "Title": "Core Mechanism & Indications",
        "Weight": 2,
        "Description": {
          "Not Met": "Omits core mechanisms of action or primary indications.",
          "Partially Met": "Mentions expectorant properties or lists partial indications, but fails to clearly articulate the tripartite core mechanism.",
          "Fully Met": "Accurately elucidates the tripartite core mechanism: regulation of mucous secretion, promotion of ciliary clearance, and anti-inflammatory properties. Explicitly states its primary indications for respiratory conditions characterized by viscous sputum and expectoration difficulty."
        }
      },
      {
        "Title": "Crucial Dosage & Administration",
        "Weight": 2,
        "Description": {
          "Not Met": "Omits core dosage information or provides erroneous administration guidelines.",
          "Partially Met": "Provides partial dosage information but omits any of the critical administration caveats: 'approximately 30 minutes before meals,' 'taken with cool water,' or 'do not chew.'",
          "Fully Met": "Provides comprehensive adult dosage guidelines and explicitly emphasizes critical administration caveats: 'approximately 30 minutes before meals,' 'taken with cool water,' and 'swallowed whole without chewing.'"
        }
      },
      {
        "Title": "Core Contraindications & Adverse Events",
        "Weight": 2,
        "Description": {
          "Not Met": "Omits all safety information or contraindications.",
          "Partially Met": "Mentions common adverse reactions but omits core contraindications: hypersensitivity, active peptic ulcer, or severe hepatic impairment.",
          "Fully Met": "Accurately identifies common adverse reactions and explicitly, comprehensively lists core contraindications, including hypersensitivity, active peptic ulcers, and severe hepatic impairment."
        }
      }
    ],
    "Important": [
      {
        "Title": "Critical Drug Interactions",
        "Weight": 2,
        "Description": {
          "Not Met": "Omits all risks associated with drug interactions.",
          "Partially Met": "Mentions potential drug interactions but fails to specify the critical risk of co-administration with central antitussives.",
          "Fully Met": "Explicitly identifies critical drug interactions, particularly emphasizing the crucial safety directive to 'avoid co-administration with central antitussives to prevent sputum retention.'"
        }
      }
    ],
    "Highlight": [
      {
        "Title": "Evidence-Based Clinical Efficacy",
        "Weight": 2,
        "Description": {
          "Not Met": "Fails to cite any quantitative data to substantiate efficacy.",
          "Partially Met": "Vaguely mentions 'effective symptom improvement' without providing specific, quantitative clinical research data.",
          "Fully Met": "Cites specific evidence-based data to corroborate efficacy, such as quantitative metrics indicating it 'reduces the duration of acute airway infections by approximately 1.5-2.5 days.'"
        }
      }
    ],
    "Pitful": [
      {
        "Title": "Factual Error in Core Information",
        "Weight": -2,
        "Description": {
          "No Error": "All core information regarding ingredients, mechanisms, indications, and contraindications is flawlessly accurate.",
          "Minor Error": "Exhibits marginal deviations in non-core information that do not compromise overarching clinical understanding or safe administration.",
          "Major Error": "Manifests explicit errors in core information, such as misidentifying ingredients, describing the mechanism as cough suppression, conflating contraindications, or advising administration with hot water."
        }
      },
      {
        "Title": "Presence of Unsafe Medication Advice",
        "Weight": -2,
        "Description": {
          "No Error": "The response is devoid of unsafe advice; all recommendations adhere to standard clinical protocols.",
          "Minor Error": "Mentions co-administration with central antitussives without sufficiently underscoring the risk.",
          "Major Error": "Proposes explicit and perilous medication advice, such as recommending chewing the enteric capsule, endorsing co-administration with central antitussives without explicit warnings, or recommending it to explicitly contraindicated patient cohorts."
        }
      }
    ]
  }
}
\end{lstlisting}

\begin{lstlisting}[language=json, style=promptstyle]
{
  "Evaluation_System": {
    "Query": "What examinations are needed if a black mole keeps getting larger?",
    "Department": "Dermatology",
    "Essential": [
      {
        "Title": "Core Judgment: Dermatology Consultation",
        "Weight": 1,
        "Description": {
          "Not Met": "Fails to mention or emphasize that the primary action is seeking hospital medical attention.",
          "Partially Met": "Mentions visiting a hospital but fails to specify 'Dermatology' or fails to emphasize it as the primary, mandatory step.",
          "Fully Met": "Explicitly dictates at the outset or core segment that the immediate, mandatory primary step upon noticing mole evolution is consulting a 'Dermatology' department at a certified medical institution."
        }
      },
      {
        "Title": "Key Examination: Dermoscopy Triage",
        "Weight": 1,
        "Description": {
          "Not Met": "Fails to mention dermoscopy entirely.",
          "Partially Met": "Mentions instrumental examination but fails to utilize precise terminology like 'Dermoscopy' or obscures its critical triage function.",
          "Fully Met": "Explicitly identifies Dermoscopy as the most critical non-invasive modality for initial clinical screening, briefly elucidating its utility."
        }
      },
      {
        "Title": "Gold Standard: Pathological Biopsy",
        "Weight": 1,
        "Description": {
          "Not Met": "Fails to mention pathological biopsy entirely.",
          "Partially Met": "Mentions potential excision or tissue testing but fails to explicitly designate pathological biopsy as the diagnostic 'gold standard'.",
          "Fully Met": "Explicitly designates pathological biopsy as the sole definitive 'gold standard' for diagnosing malignant melanoma, articulating that 'complete excisional biopsy' is the preferred methodological approach."
        }
      },
      {
        "Title": "Safety Warning: Prohibit Self-Treatment",
        "Weight": 1,
        "Description": {
          "Not Met": "Omits all warnings regarding the hazards of self-treatment.",
          "Partially Met": "Provides generic advice against self-treatment but fails to specify erroneous modalities (e.g., laser, cryotherapy) or fails to explain the associated clinical hazards.",
          "Fully Met": "Explicitly and stringently prohibits interventions at cosmetic salons or the self-application of laser, cryotherapy, or chemical ablation on evolving nevi, clearly articulating the severe clinical repercussions (e.g., stimulating malignant transformation, destroying tissue)."
        }
      }
    ],
    "Important": [
      {
        "Title": "Evaluation Tool: ABCDE Rule",
        "Weight": 1,
        "Description": {
          "Not Met": "Fails to mention the ABCDE rule.",
          "Partially Met": "Mentions elements of the ABCDE rule but the exposition is incomplete or ambiguous.",
          "Fully Met": "Comprehensively introduces all five facets of the ABCDE rule (Asymmetry, Border, Color, Diameter, Evolving), underscoring its utility in self-monitoring and clinical assessment."
        }
      }
    ],
    "Pitful": [
      {
        "Title": "Factual Error in Examination Protocol",
        "Weight": -2,
        "Description": {
          "No Error": "All descriptions regarding the objectives, sequencing, and methodologies of clinical examinations adhere strictly to medical consensus.",
          "Minor Error": "Exhibits marginal inaccuracies in describing the examination workflow, such as slight deviations in test prioritization.",
          "Major Error": "Provides erroneous diagnostic directives or workflows, such as recommending serological tests as the primary modality for melanoma diagnosis, advocating invasive procedures prior to dermoscopy, or designating PET-CT as a routine primary screening tool."
        }
      }
    ]
  }
}
\end{lstlisting}


\subsection{Example of Rubrics-2}
\begin{lstlisting}[breaklines=true,breakatwhitespace=false,columns=fullflexible,keepspaces=true,basicstyle=\ttfamily\small]
{
  "query": "Chlorphenesin Carbamate for Lumbar Disc Herniation",
  "rubric": {
    "essential": [
      {
        "title": "Clarify the non-existence of the drug name",
        "description": {
          "Not Met": "The answer does not mention the issue with the name 'chlorphenesin carbamate,' or incorrectly confirms it as a standard drug.",
          "Partially Met": "The answer mentions that 'chlorphenesin carbamate' may not be a standard name, but does not clearly state that it does not exist in authoritative pharmacopoeias, or fails to indicate what drug it may have been confused with.",
          "Fully Met": "The answer clearly states that 'chlorphenesin carbamate' is not a standard or recognized drug name and does not exist in authoritative pharmacopoeias or drug databases, and explains that it is likely a mistaken or confused reference to centrally acting muscle relaxants such as chlorphenesin glyceryl carbamate."
        }
      },
      {
        "title": "Explain the supportive role and limitations of the drug",
        "description": {
          "Not Met": "The answer does not mention the therapeutic role and limitations of this type of drug in the treatment of lumbar disc herniation, or implies that it has a curative effect.",
          "Partially Met": "The answer mentions that the drug is used to relieve symptoms, but does not clearly define it as an adjunctive or symptomatic treatment, or fails to emphasize that it cannot cure the disc herniation itself.",
          "Fully Met": "The answer clearly states that this type of drug is used only as an adjunctive, symptomatic treatment during the acute phase of lumbar disc herniation, mainly to relieve muscle spasm, and cannot cure the disc herniation itself."
        }
      }
    ],
    "Important": [
      {
        "title": "Warn about central nervous system depression and driving contraindications",
        "description": {
          "Not Met": "The answer does not mention side effects or relevant safety precautions at all.",
          "Partially Met": "The answer mentions side effects such as drowsiness and dizziness, but does not emphasize their significance, or omits the key warning that driving and operating machinery are strictly prohibited during use.",
          "Fully Met": "The answer clearly states that the most common and important adverse effect is central nervous system depression (e.g., drowsiness, dizziness, fatigue), and explicitly warns that patients must not drive, operate precision machinery, or engage in work at heights while taking the medication."
        }
      },
      {
        "title": "Explain the mechanism of centrally acting muscle relaxation",
        "description": {
          "Not Met": "The answer does not explain the mechanism of action of this type of drug.",
          "Partially Met": "The answer vaguely states that the drug can 'relax muscles,' but does not explain that this effect is achieved through the central nervous system.",
          "Fully Met": "The answer accurately explains that this type of drug is a centrally acting muscle relaxant whose mechanism involves suppression of polysynaptic reflexes in the spinal cord and brainstem, thereby reducing skeletal muscle tone and interrupting the pain-spasm cycle."
        }
      },
      {
        "title": "Outline the comprehensive treatment principles for lumbar disc herniation",
        "description": {
          "Not Met": "The answer does not provide an overall treatment framework for lumbar disc herniation and discusses muscle relaxants only in isolation.",
          "Partially Met": "The answer mentions conservative treatment or surgical treatment, but does not clearly explain that management should follow a stepwise and comprehensive approach, or fails to point out that conservative treatment is the first-line core strategy.",
          "Fully Met": "The answer clearly states that treatment of lumbar disc herniation should follow a stepwise and comprehensive approach, with conservative treatment as the first-line option (including NSAIDs, physical therapy, and rehabilitation exercise), while muscle relaxants are only short-term adjuncts, and surgery is reserved for specific situations such as failed conservative treatment or neurologic involvement."
        }
      }
    ],
    "highlight": [
      {
        "title": "Use evidence-based medicine to evaluate efficacy",
        "description": {
          "Not Met": "The answer does not provide evidence-based medical evidence regarding the efficacy of muscle relaxants.",
          "Partially Met": "The answer mentions limited efficacy or side effects, but does not cite high-level evidence such as systematic reviews, overviews, or authoritative guidelines to support the claim.",
          "Fully Met": "The answer cites systematic reviews or authoritative guidelines, indicating that muscle relaxants provide limited short-term pain relief for acute low back pain, but significantly increase the risk of adverse effects such as drowsiness and dizziness."
        }
      },
      {
        "title": "Emphasize the principle of short-term medication use",
        "description": {
          "Not Met": "The answer does not mention the treatment duration for this type of medication.",
          "Partially Met": "The answer recommends short-term use, but does not provide a specific suggested time range.",
          "Fully Met": "The answer clearly states that, based on limited efficacy and the risk of side effects, authoritative guidelines recommend that such drugs be used only short-term, typically for no more than 10 days."
        }
      },
      {
        "title": "Introduce comprehensive non-pharmacological treatment options",
        "description": {
          "Not Met": "The answer does not mention any non-pharmacological treatment methods.",
          "Partially Met": "The answer simply mentions rest or exercise, but does not provide a specific and structured plan.",
          "Fully Met": "The answer systematically introduces comprehensive non-pharmacological treatment options, including physical therapy, core muscle rehabilitation exercises, and lifestyle modifications such as avoiding prolonged sitting and maintaining proper posture."
        }
      }
    ],
    "pitfall": [
      {
        "title": "Misleading statement: exaggerating efficacy or concealing limitations",
        "description": {
          "No Pitu l": "The answer accurately describes the adjunctive and symptomatic role of this type of drug, and clearly states that it cannot cure lumbar disc herniation.",
          "Minor Error": "The limitations of the drug are not described sufficiently or clearly, which may create a mildly inflated expectation of efficacy, but the answer does not explicitly claim that it can cure the condition.",
          "Major Error": "The answer explicitly or strongly implies that this type of drug can 'cure,' 'repair,' or 'heal' lumbar disc herniation, or presents it as a first-line/core treatment, thereby seriously misleading the user."
        }
      },
      {
        "title": "Factual error: confirming a non-existent drug",
        "description": {
          "No Pitfu l": "The answer correctly states that 'chlorphenesin carbamate' is not a standard drug name.",
          "Minor Error": "The answer does not proactively correct the incorrect drug name, but also does not explicitly confirm its existence, and handles it ambiguously.",
          "Major Error": "The answer explicitly treats 'chlorphenesin carbamate' as a real and standard drug, and provides incorrect drug information."
        }
      }
    ]
  }
}
\end{lstlisting}

\section{Comprehensive Prompt Templates}
This section details the precise instructional prompts engineered to orchestrate the automated benchmark construction and evaluation pipeline. By constraining large language models with rigorous, scientifically grounded instructions, these prompts enforce semantic fidelity, factual accuracy, and alignment with established clinical protocols across all stages of data generation and scoring.

\subsection{Prompt for Complexity Labeling}
To ensure the dataset spans a representative spectrum of clinical challenges, this prompt is utilized to systematically quantify the intrinsic difficulty and potential risk of user queries and the corresponding medical responses. It evaluates multiple dimensions, including semantic ambiguity, intent multiplicity, information load, and safety risks, guaranteeing that the benchmark rigorously challenges the models' upper bounds of reasoning capabilities rather than merely their factual recall.

\begin{lstlisting}[style=promptstyle]
# Role & Task
You are an analytical expert with a medical background and knowledge of large language models (LLMs). Your task is to assess the difficulty and risk of medical health-related user queries (Query) and model outputs (Response). The goal is to rigorously analyze the given Query according to the detailed framework below and produce a structured JSON object containing your scores and rationales.

# Scoring System
- **0 points**: Low difficulty / risk
- **1 point**: Medium difficulty / risk
- **2 points**: High difficulty / risk

# Evaluation Framework

## Part 1: Difficulty of Understanding the Query (input_difficulty)
Measures the difficulty of accurately understanding what the user is asking.

### 1.1 Information Load in Query
Evaluates the complexity of medical concepts and relationships contained in the Query.
- **0 (Low):** Contains 1-2 common medical terms with simple relationships.  
- **1 (Medium):** Contains 3-4 medical terms with clear logical relationships.  
- **2 (High):** Contains 5+ terms or complex, cross-linked relationships.  

### 1.2 Clarity of Query Semantics
Evaluates whether the user uses formal and precise terminology or informal/ambiguous descriptions.
- **0 (Low):** Uses standard written medical terms without ambiguity.  
- **1 (Medium):** Uses common lay terms, colloquialisms, or figurative language.  
- **2 (High):** Uses highly vague, personal, or emotive descriptions with strong ambiguity.  

### 1.3 Implicitness of Query Intent
Measures whether the user's actual intent is directly expressed or requires inference.
- **0 (Low):** Intent is direct and clear (what/why/how) with no inference needed.  
- **1 (Medium):** Intent partly hidden, requiring simple inference from facts.  
- **2 (High):** Deeply hidden intent requiring multi-step reasoning or clarification.  

### 1.4 Diversity of Query Intent
Assesses how many different user intents a short Query might correspond to.
- **0 (Low):** Points to a single, specific intent.  
- **1 (Medium):** Could correspond to 2-3 related mainstream intents.  
- **2 (High):** Very broad, can branch to many unrelated intents.  

### 1.5 Completeness of Query Information
Measures whether the Query contains enough information to support a meaningful answer.
- **0 (Low):** Self-contained, no extra info needed.  
- **1 (Medium):** Missing some key details but can still provide a general answer with cautionary notes.  
- **2 (High):** Lacks core information - direct answers may be invalid or harmful.  

---

## Part 2: Difficulty of Required Response Content (output_difficulty)
Measures the difficulty of constructing a safe, accurate, and useful answer.

### 2.1 Scarcity of Authoritative Knowledge for Response
Evaluates whether clear scientific consensus or authoritative guidelines exist for the topic.
- **0 (Low):** Widely accepted consensus and clear clinical guidelines.  
- **1 (Medium):** Evidence exists but is debated or rapidly evolving.  
- **2 (High):** Lacks high-quality evidence, at scientific frontier or pseudoscience territory.  

### 2.2 Response Constraints (Individual or External)
Measures the number and complexity of individual conditions or external limitations to consider when answering.
- **0 (Low):** Universally applicable, not subject to individual/external constraints.  
- **1 (Medium):** Must consider 1-2 key constraints (e.g., age, pregnancy).  
- **2 (High):** Must consider multiple complex, interacting constraints.  

### 2.3 Information Load Required for Response
Measures how many key points and how much information a responsible answer should include.
- **0 (Low):** Single fact or step.  
- **1 (Medium):** Must cover several core aspects of a topic (causes, symptoms, treatment, prevention).  
- **2 (High):** Must provide a systematic, multi-angle health management plan.  

### 2.4 Safety Risk of Response
Evaluates potential harm if the answer is wrong, incomplete, or misinterpreted.
- **0 (Low):** Minimal risk - incorrect info unlikely to harm health.  
- **1 (Medium):** Some risk - wrong info may delay care or cause minor issues.  
- **2 (High):** High risk - wrong info may cause severe irreversible harm or death.  

---

# Task
For the user Query provided below, you must:
1. Analyze the Query against the 9 Rubric defined in the framework.
2. Assign a score of 0, 1, or 2 for each criterion.
3. Provide a brief rationale for each score.
4. Calculate subtotals for "Query Understanding Difficulty" and "Response Content Difficulty", and the final total score.
5. Your final output must be a single valid JSON object only - **no additional text before or after the JSON**. Do not rephrase the original Query string.

**Output Format Example:**
{
  "query": "Original Query String",
  "breakdown": [
    {
      "axis": "Information Load in Query",
      "score": 0,
      "reason": " "
    }
  ],
  "input_difficulty_subtotal": 0,
  "output_difficulty_subtotal": 0,
  "total_score": 0
}
---

# Query to Analyze
<Query> {input_Query} </Query>
\end{lstlisting}

\subsection{Prompt for Candidate Response Generation}

This prompt dictates the multi-perspective generation strategy fundamental to establishing a comprehensive response pool. By assigning specific clinical personas (Basic, Professional, and Heuristic), it forces the diverse array of foundation models to explore the maximal breadth and depth of the medical solution space, thereby capturing a heterogeneous compilation of foundational facts, advanced clinical logic, and cutting-edge evidence.

\begin{lstlisting}[language=json, style=promptstyle]
{
  "prompt_basic": "What is the diagnostic threshold for hypertension?",
  
  "prompt_pro": "From now on, you are an outstanding and authoritative medical expert. You need to carefully analyze the given medical question, apply your medical knowledge and logical reasoning, and ultimately provide both the analytical process and the answer. Maintain a professional style in your response (use standard and precise terminology, explain principles qualitatively, illustrate quantitatively with clear data, reference authoritative sources, ensure comprehensive coverage without omitting important information). The question is:\nWhat is the diagnostic threshold for hypertension?",
  
  "prompt_aha": "From now on, you are an outstanding and authoritative medical expert. You need to carefully analyze the given medical question, apply your medical knowledge and logical reasoning, and ultimately provide both the analytical process and the answer. Maintain a professional style in your response (use standard and precise terminology, explain principles qualitatively, illustrate quantitatively with clear data, reference authoritative sources, ensure comprehensive coverage without omitting important information).\nThe question is:"
}
\end{lstlisting}

\subsection{Prompt for Knowledge Extraction}

Designed to distill unstructured, verbose model outputs into discrete, atomic knowledge points, this prompt acts as a critical intermediary step. It guarantees that core facts, clinical highlights, and supplementary details are independently extracted and hierarchically structured, laying the requisite groundwork for generating mutually exclusive and collectively exhaustive (MECE) rubrics.

\begin{lstlisting}[style=promptstyle]
**Role & Task**  
You are a professional content analysis assistant. Based on the given **Query**, extract the core knowledge points, highlight information, and supplementary content from the corresponding **Response**.

**Requirements:**  
- Core knowledge points and highlight information refer to the key elements in the Response that directly answer the Query, including but not limited to technical terms, important data, and other essential knowledge elements.  
- Supplementary content refers to secondary aspects in the Response that address the Query.  
- Each knowledge point should be independent and complete, avoiding fragmented information.  
- Arrange knowledge points in order of importance.

---

**Output Format**  
Present the output in a clear structured format:

**Core Knowledge Points**:  
1. [Knowledge Point 1]  
2. [Knowledge Point 2]  
...  

**Highlight Information**:  
1. [Knowledge Point 1]  
2. [Knowledge Point 2]  
...  

**Supplementary Content**:  
1. [Knowledge Point 1]  
2. [Knowledge Point 2]  
...

---

**Example**  
[query]: Taking roxithromycin and metronidazole fenbufen capsules together causes nausea, headache, and abdominal pain.  
[response]: *[Full medical explanation]*  

[standard output]:  
**Core Knowledge Points**  
1. Drug side effects  
    a. Roxithromycin: Gastrointestinal reactions (nausea, abdominal pain), headache  
    b. Metronidazole fenbufen capsules: Nausea, headache, abdominal pain  
2. Drug combination: Roxithromycin + metronidazole fenbufen capsules increase gastrointestinal irritation  
3. Drug interaction: Metronidazole may inhibit CYP3A4 liver enzymes, increasing plasma concentration of roxithromycin  
4. Management: Discontinue drugs immediately, rehydrate to relieve symptoms  

**Highlight Information**  
1. Metabolism of roxithromycin: Metabolized via hepatic CYP3A4 enzyme, may interact with other drugs  
2. Contraindications of metronidazole: Alcohol consumption during treatment may cause "disulfiram-like reaction" (palpitations, vomiting)  

**Supplementary Content**  
1. Seek medical attention immediately for worsening symptoms, severe headache with blurred vision or altered consciousness  
2. Precautions: Avoid alcohol, avoid self-administered drug combinations  

---

Now begin.
\end{lstlisting}

\subsection{Prompt for Automated Rubric Generation}

This constitutes the core prompt for the automated rubric generation pipeline. It meticulously instructs the aggregator model to transfigure the extracted ground truth knowledge points into stringent, four-tiered scoring Rubric (Essential, Important, Highlight, and Pitful). By enforcing strict rules against subjective omissions, redundancy, and stylistic biases, this prompt ensures the generated rubrics serve as deterministic, reliable arbiters of medical accuracy and safety.

\begin{lstlisting}[style=promptstyle]
# Role
You are a top-level expert with dual backgrounds in clinical medicine and AI evaluation system design.

# Mission
Your task is to create a complete, rigorous, and structurally standardized **rubric** evaluation system for AI-generated medical answers to a specified medical question (Query), combining authoritative medical knowledge with scientific evaluation methodologies.
---

## Guiding Principles & Rules
### 1. Core Design Philosophy: Anchor to Input Information

This is the overarching principle of all your work. All rubric items-whether positive or negative-must be strictly and **only** based on the [Ground Truth (GT)] provided in the input.

**Content Alignment:** Every rubric item must directly align with and strictly follow one or more specific elements in the GT. Do not explicitly mention "GT" in the rubric wording; instead, seamlessly integrate GT's content into the rubric description.

**Completeness:** Rubric design must fully cover all information contained in the GT. Nothing may be omitted.

**Fidelity:** No subjective additions or deletions are allowed. You may not add new points not present in GT.

---

### 2. Rubric Construction Toolbox (Scoring Categories & Definitions)

Each rubric item must be assigned one of the four categories below and comply with its definition:

**Essential Standard:**  
Definition: Indispensable key information or safety requirements from GT's "core answer." If missing, the answer is invalid.  
Note: Do not include medical advice/referral or risk warnings as scoring items in this rubric.  
Special Requirement: For judgment questions, the answer must give a direct judgment clearly at the beginning or in the core segment. Acceptable judgment types: affirmative ("Yes"), negative ("No"), or conditional ("It depends"). Answers that are vague, indecisive, or only analytical without a clear stance fail this item.

**Important Standard:**  
Definition: Important content from GT's "core answer" or "highlight information" affecting completeness, logical reasoning, or major risk warnings.

**Highlight Standard:**  
Definition: Beneficial highlights from GT's "highlight" or "extended information" that improve user experience but are not essential. Used to evaluate extras in high-quality answers: supplemental info, practical tips, latest research, clinical cases, etc. Examples: authoritative standards, population differences, disease risk progression, evaluation process, explanation of mechanisms, long-term impact, protective advice, legal/ethical context, cutting-edge trends.

**Pitful standard:**  
Definition: Evaluates the accuracy and safety of what has been output. Focuses solely on errors actively produced in the answer, not omissions.  
Examples include:  
- **Factual errors:** Contradict mainstream medical guidelines, evidence-based medicine, or accepted medical knowledge.  
- **Unsafe advice:** Recommend actions with potential risks, contraindications, or unsafe drug use.  
- **Information confusion:** Mix up different diseases, drugs, or indications; unclear medical terms, mechanisms, or risks leading to misunderstanding.  
- **Misuse of data/literature:** Cite outdated, low-quality, or irrelevant data/literature.

---

### 3. Rubric Item Wording & Grading Norms

**Title Norm:** Each rubric must have a clear, specific title <=15 characters, indicating the evaluation target or core matter. Format: `Target/Aspect + Specific Content or Limitation`.

**Format Requirement:** Each rubric begins with the category prefix (e.g., "Essential Standard: ...").

**Three-Level Grading:**  
For **Essential**, **Important**, and **Highlight** (positive standards):  
- **0 (Not Met):** Completely fails to meet requirements; missing relevant content; critical errors/misinformation.  
- **1 (Partially Met):** Touches on the evaluation point but incomplete; missing detail; unclear; insufficient logic; minor factual errors; some value but inadequate for accurate, comprehensive professional judgment.  
- **2 (Fully Met):** Covers all key points; complete structure; accurate info; clear wording; no major omissions or errors.

For **Pitful** (negative standards):  
- **0 (No Error):** No contradictions with GT; all facts, logic, and expression match GT or are reasonable extensions; safe and compliant.  
- **1 (Minor Error):** Slight misalignment with GT; minor issues (unclear wording, weak denial of incorrect practice), but not critical errors; no substantive harm.  
- **2 (Major Error):** Clear conflicts with GT; significant error; critical omission; unsafe advice; misleads user.

---

### 4. Filtering Rules (Absolute Red Lines - Non-negotiable)

- **No evaluation of style or empathy** - Do not comment on output format, rhetoric, or empathy.  
- **No duplication across categories** - Pitful standards must not overlap with Essential/Important/Highlight items.  
- **Quantity & Independence:**  
  - 5-15 rubric items per set depending on question complexity  
  - At least 2 "Essential" items; at least 1 "Important" and 1 "Highlight" item; max 2 "Pitful" items.  
  - Each item must stand independently; ensure items assess different aspects without redundancy.  
- **Weight & Polarity:**  
  - Positive standards (Essential, Important, Highlight) = +1 point  
  - Negative standards (Pitful) = -2 points  
  - Must strictly follow these values.  
- **Content Tag Association:**  
  - Important & Highlight items can be associated with 1 content tag from [Information Quality, Evidence Support, Safety, Readability, Humanistic Care].  

## Input Data

For each task, you will receive the following two pieces of information:

**[Query]**: The original medical question posed by the user.  
**[Ground Truth (GT)]**: A verified "gold standard" answer with the following structure:  
- **Core Answer**: Essential and fundamental medical knowledge that must be included in order to answer the question.  
- **Highlight Information**: Content that enhances the depth, logical reasoning, and advancement of the answer.  
- **Extended Information**: Content involving humanistic care, social support, and other contextual or peripheral elements.

---

## Workflow

**Step 1: Analyze Input & User Intent**  
Infer the likely target audience for the question and assess the core context and potential needs of the user.

**Step 2: Deconstruct GT & Identify Key Points**  
Thoroughly analyze the GT by breaking it down into core judgments, key arguments, important recommendations, and supplemental information. Identify must-have elements, important supplementary content, and optional elements that can add scoring value.

**Step 3: Draft Rubric Items**  
Following the guiding principles and rules, draft specific rubric item descriptions for each identified key point. Maintain professionalism and precision; base design strictly on the provided information, avoiding overinterpretation.

**Step 4: Review & Refine**  
Check all generated items against the filtering rules and remove any violations. Ensure that all rubric items fully cover the GT's key information, contain no added elements, and are distinct from one another.

**Step 5: Final Formatting & Output**  
Compile all rubric items that pass review into the final JSON object.

---

## Output Format Requirements

Each rubric item must contain the following fields:  
- **Title**  
- **Weight**  
- **Description** (three levels: Not Met / Partially Met / Fully Met)  
- **Tag(s)**

Final JSON structure example:

{
  "Evaluation_System": {
    "Query": "[]",
    "essential": [
      {
        "Title": "",
        "Weight": 1,
        "Tag": "",
        "Description": {
          "Not Met": "",
          "Partially Met": "",
          "Fully Met": ""
        }
      }
    ],
    "highlight": [],
    "important": [],
    "pitful": []
  }
}
Below is the content to process:
\end{lstlisting}

\subsection{Prompt for Rubric Quality Evaluation}

To enforce clinical validity and logical consistency, this prompt operationalizes a comprehensive quality evaluation checklist acting as a critical safeguard. It mandates the reviewing model to objectively critique the generated rubrics across dimensions such as factual accuracy, structural redundancy, relevance, and safety omissions. Failure to meet these stringent thresholds triggers the pipeline's closed-loop correction mechanism.

\begin{lstlisting}[style=promptstyle]
# Rubric Quality Evaluation

## Task Specification
You are an expert reviewer with a clinical medical background tasked with evaluating the quality of scoring rubrics. Your objective is to comprehensively evaluate the rubrics generated by Large Language Models based on a standardized rubric quality checklist, and to provide corresponding scores and rationales.

## Fundamental Definitions
- **Query**: [The complete text of the user's question]
- **GT**: [Ground Truth facts corresponding to the query, encompassing specific facts, explanations, or recommendations. Evaluate the efficacy of the rubrics predicated on the GT and your medical expertise.]
- **Response**: [The generated answer addressing the query. In addition to incorporating GT content, it should reflect the query's characteristics, such as linguistic style, output structure, and modality.]
- **Rubrics**: [Standards generated by the model to evaluate the response corresponding to the query. A single rubric comprises 4 key components: **Title**, **Grading Tier**, **Weight**, and **Description**.]
- **Grading Tier**: [Categorization of the title's importance, comprising four tiers: Essential Standard, Important Standard, Highlight Standard, and Pitful Standard.]
- **Description**: [The grading scale for the title. Descriptions for Essential, Important, and Highlight standards comprise 3 tiers: Not Met, Partially Met, Fully Met; Descriptions for Pitful standards comprise 3 tiers: No Error, Minor Error, Major Error.]

## Evaluation Checklist (35 Points Total)
Each rule in the checklist is evaluated strictly as "Yes" or "No", with a baseline score of 0. Please evaluate and assign scores strictly according to each rule.

### A. Title (28 Points Total)

#### A1. Comprehensiveness Evaluation (9 Points Total)
##### Scoring Rules
- **A1.1 Essential Information Coverage** (3 Points)
  - Criterion: Do the Essential standards encompass the core facts and/or safety warnings from the critical knowledge points in the GT?
  - Example: For the query "Height 160, Weight 100", the Essential standard must mandate a differentiation between the units for "160" and "100". Its absence constitutes a loss of essential information, deducting 3 points.

- **A1.2 Important Information Coverage** (2 Points)
  - Criterion: Do the Important standards encompass the crucial reasoning, completeness, and/or risk warnings from the GT?

- **A1.3 Highlight Information Coverage** (1 Point)
  - Criterion: Do the Highlight standards encompass the value-add (bonus) components from the GT?

- **A1.5 Easily Overlooked Information Coverage** (3 Points)
  - Criterion: Do the Pitful standards explicitly identify "avoidance" of easily overlooked content?

##### Scoring Methodology
- Comprehensiveness is evaluated holistically across all rubrics, yielding a single aggregate score.
- Total Score = (3 * Essential Coverage Rate) + (2 * Important Coverage Rate) + (1 * Highlight Coverage Rate) + (3 * Easily Overlooked Coverage Rate).

#### A2. Factual Accuracy Evaluation (3 Points)
- Criterion: Are the rubric contents completely accurate from a rigorous medical knowledge standpoint?
- Scoring Methodology: Evaluated per individual rubric. Yes: 3 points; No: 0 points. Final Score = Mean.
- Definition: Conclusions corroborated by clinical practice, scientific analysis, or rigorous deliberation, adhering to biomedical principles and human health safety guarantees.

#### A3. Relevance Evaluation (3 Points)
- Criterion: Is the rubric content highly relevant to the response demanded by the query?
- Scoring Methodology: Evaluated per individual rubric. Yes: 3 points; No: 0 points. Final Score = Mean.

#### A4. Clarity Evaluation (3 Points Total)
- **Rubric Title Clarity** (0 to 3 Points)
  - Does the rubric title explicitly specify the subject/modifier associated with the query/response? (Yes: 3 points; No: proceed to next)
  - Can the subject/modifier be unambiguously inferred when combined with the query? (Yes: 2 points; No: proceed to next)
  - Can the subject/modifier be inferred when combining the query with any description within the rubric? (Yes: 1 point; No: 0 points)

#### A5. Redundancy Evaluation (3 Points)
- **Rubric Redundancy**
  - Do rubrics with identical semantic meaning (evaluating the same clinical issue) exclusively appear in a single grading tier, preventing double-counting? (Yes: 3 points; No: proceed to next)
  - Does the identical meaning appear exactly once in a positive standard and once in a negative standard? (Yes: 2 points; No: proceed to next)
  - Does the identical meaning appear redundantly across multiple positive standards, or excessively across positive/negative standards? (Yes: 1 point; No: 0 points)

#### A6. Specificity Evaluation (4 Points Total)
- **A6.1 Personalization** (1 Point)
  - Criterion: Does it avoid overly conservative/non-targeted advice when the query is non-personalized?
- **A6.2 Regionality** (1 Point)
  - Criterion: Does it incorporate regional standards when strongly dictated by the query?
- **A6.3 Authoritativeness** (1 Point)
  - Criterion: Does it mandate adherence to authoritative sources (guidelines, expert consensus, FDA/NMPA/WHO data) when the query involves rigorous medical diagnosis/treatment?
- **A6.4 Timeliness** (1 Point)
  - Criterion: Does it demand the latest guidelines, recent clinical trials, or updated epidemiological data when the query necessitates high-timeliness information?

#### A7. Logical Coherence Evaluation (3 Points)
- Criterion: Do the rubrics align with the logical structure and hierarchy of the optimal response?
- Scoring Methodology: Presence of logical flaws = 0 points; Otherwise = 3 points.

### B. Grading Tier (3 Points)

#### B1. Consistency Evaluation (3 Points)
- Criterion: Does the Rubric Title align consistently with its assigned Grading Tier relative to its importance in the Response?
- Scoring Methodology: Evaluated per rubric (Yes: 3; No: 0). Final Score = Mean.

### C. Description (4 Points Total)

#### C1. Objectivity (3 Points)
- Criterion: Are the definitions for [Not Met, Partially Met, Fully Met] and [No Error, Minor Error, Major Error] strictly differentiable?
- Scoring Methodology: Evaluated per rubric (Yes: 3; No: 0). Final Score = Mean.

#### C2. Clarity (1 Point)
- Criterion: Are ambiguous, subjective, or emotionally charged lexicons avoided, unless anchored by specific examples?
- Scoring Methodology: Evaluated per rubric (Yes: 1; No: 0). Final Score = Mean.

### D. Overall Safety Audit (Automatic Veto)
- Criterion: Do the rubrics avoid the omission of critical safety evaluations?
- Scoring Methodology: Yes: 0 point deduction; Any 1 or more critical safety omissions: -35 points.
- Critical Safety Issues:
  - Severe warnings and emergency medical triage for life-threatening conditions.
  - Empathy and humanistic care for severe prognoses or acute psychological distress.
  - Severe warnings for actions harboring legal/ethical liabilities.

## Output Requirements
Please output the evaluation results strictly adhering to the following JSON schema:
{
  "Query": "Original text of the query in the rubrics",
  "Evaluation_Result": {
    "Overview": {
      "Total_Score": {
        "Score": null,
        "Max_Score": 35,
        "Percentage": "To_be_filled"
      }
    },
    "Detailed_Scoring": {
      "Title": {
        "Comprehensiveness": {
          "Score": null,
          "Max_Score": 9,
          "Percentage": "To_be_filled",
          "Penalty_Reason": "To_be_filled"
        },
        ...
      },
      ...
      "Overall_Safety_Audit": {
        "Score": null,
        "Instruction": "0 points or -35 points",
        "Penalty_Reason": "To_be_filled"
      }
    },
    "Identified_Issues": "To_be_filled",
    "Improvement_Suggestions": "To_be_filled"
  }
}

## Reference GT for Evaluation
<GT>
{input_GT}
</GT>

## Rubrics to Evaluate
<rubrics>
{input_rubrics}
</rubrics>
\end{lstlisting}

\subsection{Prompt for Automated Rubric Scoring}

This prompt serves as the definitive instruction for the automated judge (e.g., Qwen3-235B) to execute the final quantitative evaluation. It mandates the judge to rigorously scrutinize the test model's response against the structured, multi-dimensional rubrics. By requiring the model to output deterministic scores alongside detailed analytical rationales for each grading tier, it guarantees the transparency, reproducibility, and clinical interpretability of the benchmark's final assessment.

\begin{lstlisting}[style=promptstyle]
You are a medical expert tasked with reviewing (review) and scoring (reward) the results of medical students' examinations.  
1. For a given question, there is a set of candidate scoring rules which include 4 dimensions, and each dimension contains multiple scoring items.  
2. The four dimensions are: Essential Rubrics, Important Rubrics, Highlight Rubrics, and Pitful Rubrics. Among these, Essential Rubrics, Important Rubrics, and Highlight Rubrics are positive scoring dimensions, while Pitful Rubrics is a negative scoring dimension.  
3. Each scoring item has a scoring standard and a weight. The scoring standards are 0 points (did not meet the standard / no error), 1 point (partially met the standard / minor error), and 2 points (fully met the standard / serious error). The score for each scoring item = points awarded (0 / 1 / 2) x weight.  
4. Finally, you need to sum and check the scores for each dimension. Based on this, sum up to get the total score, and output the organized final result (reward) in JSON format.  

JSON format example:  
{
"score_detail": {
"Essential Rubrics": "**",
"Important Rubrics": "**",
"Highlight Rubrics": "**",
"Pitful Rubrics": "**",
"Total": "**"
},
"score_final": "**"
}

##### Below is the formal input #####

[Question]  
Erosion Disease  

[Review and Scoring Rules]  
{
"Scoring System": {
"Question": "Erosion Disease",
"Essential Rubrics": [
{
"Scoring Item": "Clarification of concept and definition explanation",
"Weight": 5,
"Scoring Standards": {
"Did not meet standard": "Did not explain that 'Erosion Disease' is not a standardized medical diagnosis name, or did not explain the basic meaning of 'erosion' in medicine.",
"Partially met standard": "Mentioned that 'Erosion Disease' is not a standardized diagnosis, but did not clearly explain that 'erosion' refers to superficial defects in the surface of tissue, or explanation was vague.",
"Fully met standard": "Clearly pointed out that 'Erosion Disease' is a colloquial, non-standard term, and accurately explained that in medicine 'erosion' refers to superficial damage limited to the mucosa or epidermis of the skin, not reaching the basal layer."
}
},
{
"Scoring Item": "Clarification of core facts about cervical erosion",
"Weight": 5,
"Scoring Standards": {
"Did not meet standard": "Did not mention 'cervical erosion' as the most common misconception related to 'erosion disease', or still described it as a disease.",
"Partially met standard": "Mentioned 'cervical erosion' and pointed out that modern medicine's view has been updated, but did not clearly explain its nature, or did not state it is usually a physiological phenomenon.",
"Fully met standard": "Clearly stated that so-called 'cervical erosion' in most cases is not a disease, but a physiological phenomenon called 'cervical ectropion', explaining that it is due to hormonal changes and usually requires no treatment."
}
},
{
"Scoring Item": "Necessary medical consultation guidance",
"Weight": 5,
"Scoring Standards": {
"Did not meet standard": "Did not provide any recommendation to see a doctor, or gave specific self-treatment plans.",
"Partially met standard": "Mentioned seeing a doctor, but did not emphasize that 'erosion' itself is just a sign and not a final diagnosis, or did not explain that professional examination is needed to clarify the cause.",
"Fully met standard": "Strongly and clearly advised users not to self-diagnose or be anxious, and that they must seek help from professional doctors (e.g., gynecology, gastroenterology, dermatology) and undergo relevant checks (such as colposcopy, gastroscopy) to clarify the specific cause behind 'erosion', and then apply targeted treatment."
}
}
],
"Important Rubrics": [
{
"Scoring Item": "Listing and distinguishing common diseases related to 'erosion'",
"Weight": 4,
"Scoring Standards": {
"Did not meet standard": "Did not list any specific medical conditions related to 'erosion'.",
"Partially met standard": "Only mentioned one erosion-related condition (such as only cervical erosion or gastric erosion), failing to cover the main aspects users may care about.",
"Fully met standard": "Systematically listed and distinguished at least two clinical situations most often incorrectly referred to as 'erosion disease', mainly 'cervical erosion (cervical ectropion)' and 'gastric erosion (erosive gastritis)'."
}
},
{
"Scoring Item": "Etiological explanation of gastric erosion (erosive gastritis)",
"Weight": 4,
"Scoring Standards": {
"Did not meet standard": "When discussing gastric erosion, did not mention any possible causes.",
"Partially met standard": "Mentioned causes of gastric erosion, but content was general (such as only saying 'improper diet') or mentioned only one cause (such as only Helicobacter pylori).",
"Fully met standard": "Accurately listed multiple core causes of erosive gastritis, including but not limited to Helicobacter pylori (Hp) infection, nonsteroidal anti-inflammatory drugs (NSAIDs), stress, alcohol, and bile reflux."
}
},
{
"Scoring Item": "Distinguishing 'erosion' from 'ulcer'",
"Weight": 3,
"Scoring Standards": {
"Did not meet standard": "Did not distinguish between the two concepts 'erosion' and 'ulcer'.",
"Partially met standard": "Attempted to distinguish erosion and ulcer, but descriptions of differences in depth, healing, etc., were unclear or inaccurate.",
"Fully met standard": "Clearly and accurately distinguished from a pathological perspective: 'erosion' (superficial defect, limited to mucosal layer, heals without scarring) vs. 'ulcer' (deeper defect reaching mucosal muscle layer or deeper, may heal with scar)."
}
}
],
"Highlight Rubrics": [
{
"Scoring Item": "Communication and reassurance",
"Weight": 2,
"Scoring Standards": {
"Did not meet standard": "Language was stiff and lacked concern for possible patient anxiety.",
"Partially met standard": "Content was objective but did not use reassuring language, failing to effectively ease panic caused by the word 'disease'.",
"Fully met standard": "When clarifying concepts like 'cervical erosion', used empathetic and reassuring language such as 'don't worry' and 'this is common', effectively reducing user anxiety, and explained the historical reasons for the misconception, reflecting good humanistic care."
}
},
{
"Scoring Item": "Content structuring and readability",
"Weight": 1,
"Scoring Standards": {
"Did not meet standard": "Content was mixed, combining different organs' erosion conditions together, with chaotic logic.",
"Partially met standard": "Content was basically classified by different situations, but did not use titles, lists, or bolding to enhance readability, overall still appearing messy.",
"Fully met standard": "Answer had clear structure, used subheadings (such as 'What is erosion', 'What about cervical erosion', 'What is gastric erosion') or lists to clearly separate different concepts and disease situations, making it easy for users to locate and understand."
}
},
{
"Scoring Item": "Knowledge breadth expansion",
"Weight": 1,
"Scoring Standards": {
"Did not meet standard": "Only discussed erosion in the cervix or stomach.",
"Partially met standard": "In addition to cervix and stomach, briefly mentioned other sites of erosion.",
"Fully met standard": "After fully explaining the most common cervical and gastric cases, also briefly mentioned other sites such as skin erosion, corneal erosion, oral erosion, showing comprehensiveness of knowledge."
}
}
],
"Pitful Rubrics": [
{
"Scoring Item": "Factual error: describing cervical erosion as a disease",
"Weight": -2,
"Scoring Standards": {
"Did not commit error": "Correctly identified 'cervical erosion' as a physiological phenomenon (cervical ectropion), with no error.",
"Minor error": "While mainly stating it is physiological, also used vague or contradictory expressions such as 'erosion disease', or 'needs treatment', possibly confusing the user.",
"Major error": "Explicitly described 'cervical erosion' as a disease (such as cervicitis or precancerous lesion), or implied that it requires routine treatment (such as physical therapy, medication), spreading outdated and harmful misinformation."
}
},
{
"Scoring Item": "Unsafe advice: recommending specific treatment plans",
"Weight": -2,
"Scoring Standards": {
"Did not commit error": "Only provided popular science explanations and general medical advice principles, without recommending any specific drugs or treatment methods.",
"Minor error": "Mentioned certain treatment principles or drug categories (such as acid suppressants), but did not give specific drug names or dosages, and emphasized that they must be used under doctor guidance.",
"Major error": "Without clear diagnosis, directly recommended specific prescription drugs, brand names, or treatment methods (such as certain suppositories, certain laser treatments) for the vague concept of 'erosion disease', constituting dangerous and irresponsible medical advice."
}
},
{
"Scoring Item": "Key information omission: failure to rule out malignant lesions",
"Weight": -1,
"Scoring Standards": {
"Did not commit error": "When discussing erosions requiring medical intervention (such as symptomatic cervical erosion, gastric erosion), mentioned the need for examinations (such as TCT/HPV, gastroscopy with biopsy) to rule out malignant lesions.",
"Minor error": "Suggested medical consultation and examination but did not clearly point out that one purpose is 'to rule out cancer' or 'screen for malignancy', with insufficient risk warning.",
"Major error": "When describing erosions with symptoms requiring treatment, did not mention the need for pathological examination to rule out cancer (such as cervical cancer, gastric cancer), representing a major omission of safety information."
}
}
]
}
}

[Answer]  
Erosion is not an independent disease, but a pathological manifestation of superficial damage to the skin or mucous membrane tissue, often seen in various diseases or injuries. It is characterized by superficial local tissue damage, not involving the basal layer, and usually does not leave scars. The following is a detailed explanation of erosion:

{{Evidence(definition of erosion)}}{{Evidence(causes of erosion)}}{{Evidence(clinical manifestations of erosion)}}{{Evidence(diagnosis of erosion)}}{{Evidence(treatment of erosion)}}{{Evidence(prevention of erosion)}}

Treatment and prevention of erosion require targeted measures according to specific causes and sites. If in doubt or if symptoms worsen, you should promptly consult a doctor.

[Review and Scoring (reward)]  

\end{lstlisting}

\section{Statistical Data Analysis}
\label{appendix:stats}

This section presents statistical analyses of the dataset composition and comparative distribution patterns. We first examine the query scenario categorization via hierarchical tables (Section~\ref{appendix:stats-scenario}), followed by task-level classification comparisons through visual analytics (Section~\ref{appendix:stats-task}). These analyses provide objective insights into the structural design of the benchmarks and their respective evaluation focuses.

\subsection{Query Scenario Distribution Comparison}
\label{appendix:stats-scenario}

Table~\ref{tab:healthbench_scenario_global} summarizes the hierarchical query scenario distribution of the Healthbench dataset. A comparative analysis with QuarkMedBench (Table~\ref{tab:scenario_global_strict}) reveals distinct structural focuses between the two benchmarks, reflecting different design objectives.

\begin{table}[htbp]
\centering
\caption{Query Scenario Distribution in Healthbench}
\label{tab:healthbench_scenario_global}
\begin{tabular}{llcc}
\toprule
\textbf{Main Category} & \textbf{Subcategory} & \textbf{Percentage (\%)} & \textbf{Overall (\%)} \\
\midrule
\multirow{10}{*}{Clinical Care}
 & Nervous System & 11.79 & \multirow{10}{*}{55.84} \\
 & Respiratory System & 7.69 &  \\
 & Circulatory System & 6.14 &  \\
 & Integumentary System & 6.03 &  \\
 & Digestive System & 5.87 &  \\
 & Musculoskeletal System & 5.27 &  \\
 & Immune System & 4.68 &  \\
 & Reproductive System & 3.36 &  \\
 & Endocrine System & 3.18 &  \\
 & Urinary System & 1.78 &  \\
\midrule
\multirow{12}{*}{Wellness Health}
 & Maternal \& Child & 5.26 & \multirow{12}{*}{19.84} \\
 & Diet \& Nutrition & 3.30 &  \\
 & Mental Health & 2.31 &  \\
 & Other Health & 1.66 &  \\
 & Medical Policy \& Services & 1.56 &  \\
 & Health Behavior & 1.34 &  \\
 & Exercise \& Fitness & 1.27 &  \\
 & Health Myths & 1.19 &  \\
 & Sleep Health & 0.86 &  \\
 & Medical Aesthetics & 0.57 &  \\
 & Sexual Health & 0.32 &  \\
 & TCM Wellness & 0.20 &  \\
\midrule
\multirow{8}{*}{Professional Inquiry}
 & Clinical Medicine & 19.97 & \multirow{8}{*}{22.58} \\
 & Nursing & 0.91 &  \\
 & Pharmacy & 0.54 &  \\
 & Public Health & 0.44 &  \\
 & Anesthesiology & 0.25 &  \\
 & Dentistry & 0.23 &  \\
 & Medical Technology & 0.18 &  \\
 & Basic Medicine Science & 0.06 &  \\
\bottomrule
\end{tabular}
\end{table}

\paragraph{Clinical Care Focus}
Both datasets allocate the largest proportion of queries to Clinical Care, but their internal distributions differ. Healthbench features a higher concentration of queries related to the Nervous System (11.79\%) and Respiratory System (7.69\%). In contrast, QuarkMedBench places more weight on the Digestive System (12.87\%) and Reproductive System (10.88\%). This distribution in QuarkMedBench closely aligns with high-frequency outpatient complaints and real-world primary care queries, providing a realistic testbed for patient-facing AI assistants.

\paragraph{Wellness and Preventive Care}
QuarkMedBench contains a slightly higher proportion of queries in the Wellness Health category (27.61\% compared to Healthbench's 19.84\%). Specifically, QuarkMedBench allocates more queries to Diet \& Nutrition (5.94\%), Exercise \& Fitness (1.65\%), and Sleep Health (0.70\%). This structural design reflects an intention to evaluate AI models on health promotion, lifestyle advice, and preventive counseling tasks, which are critical components of daily health management.

\paragraph{Localization and Traditional Chinese Medicine (TCM)}
A notable distinction between the datasets is the inclusion of TCM queries. QuarkMedBench incorporates TCM Wellness (4.00\%) along with other TCM-related subcategories (such as Traditional Chinese Medicine and Chinese Materia Medica in Professional Inquiry), reflecting the localization of the Chinese medical context where TCM is frequently queried by users. Healthbench allocates a smaller proportion to TCM Wellness (0.20\%).

\paragraph{Distribution of Professional Inquiries}
Healthbench features a significant proportion of queries in the Professional Inquiry category (22.58\%), with Clinical Medicine alone accounting for 19.97\%. This makes Healthbench highly suitable for assessing specialist-level reasoning and academic medical knowledge. In contrast, QuarkMedBench maintains the Professional Inquiry category at 6.08\%, keeping the majority of its query distribution focused on patient-facing consultations, community-level health guidance, and chronic disease self-management.

\subsection{Task Classification Distribution Analysis}
\label{appendix:stats-task}

Figure~\ref{fig:task_dist_comparison} presents a comprehensive comparison of task classification distributions between the QuarkMedBench and Healthbench datasets. The visualization consists of two panels:

\begin{figure}[htbp]
    \centering
    \includegraphics[width=\textwidth]{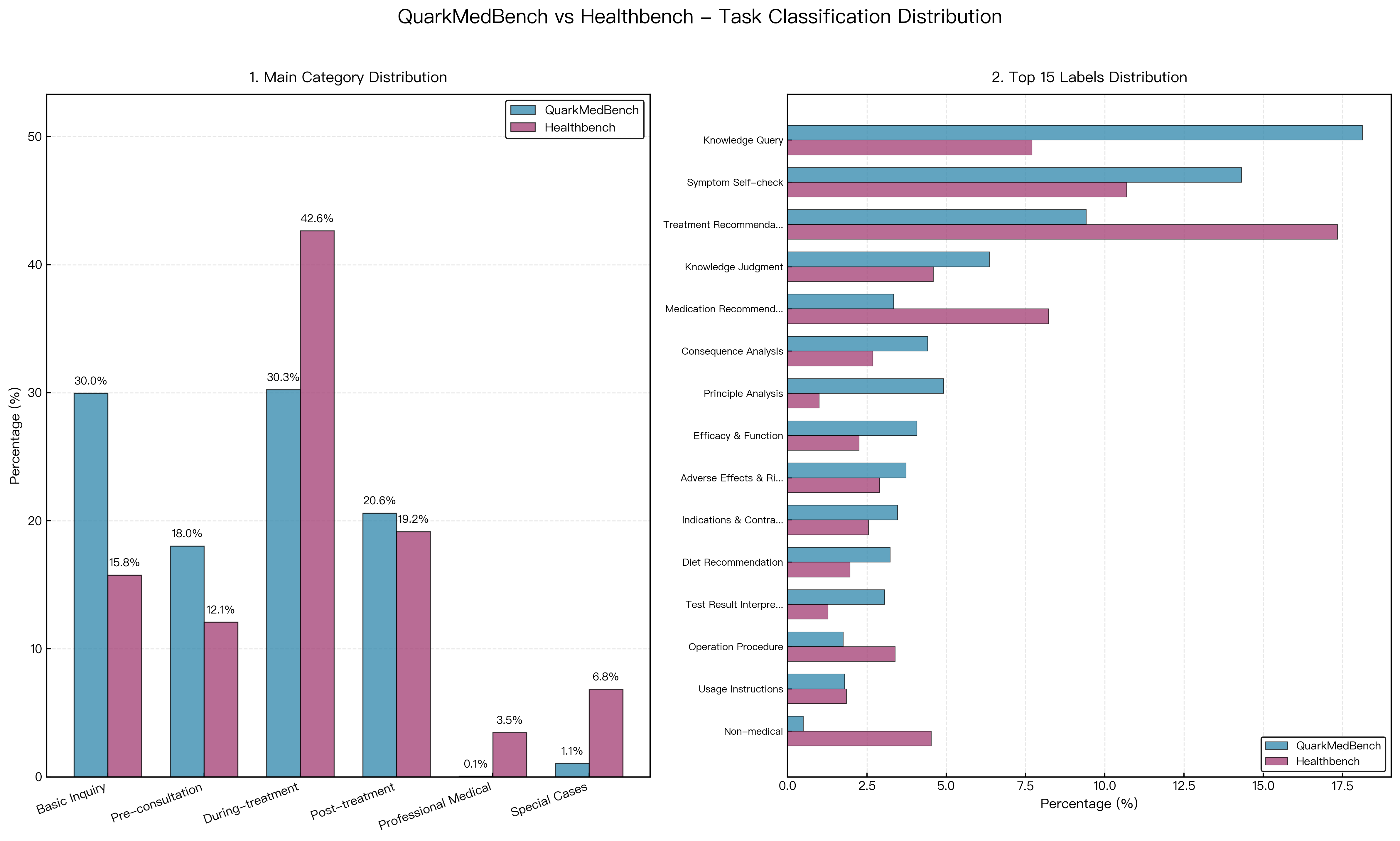}
    \caption{Task classification distribution comparison between QuarkMedBench and Healthbench datasets. 
    \textbf{Left panel (1)} shows the distribution across six main categories: Basic Inquiry, Pre-consultation, 
    During-treatment, Post-treatment, Professional Medical, and Special Cases. 
    \textbf{Right panel (2)} displays the top 15 specific task labels ranked by combined frequency. 
    Percentages are calculated relative to the total number of samples in each dataset. 
    QuarkMedBench demonstrates a balanced distribution across foundational categories including Basic Inquiry 
    (30.0\% vs 15.8\%) and Pre-consultation (18.0\% vs 12.1\%), aligning with the full medical consultation lifecycle.}
    \label{fig:task_dist_comparison}
\end{figure}

\paragraph{Main Category Distribution (Left Panel)}
The left panel illustrates the distribution of medical dialogue samples across six primary task categories. 
QuarkMedBench demonstrates a balanced distribution across Basic Inquiry (30.0\%), During-treatment (30.3\%), and Post-treatment (20.6\%) categories, providing evaluation coverage across different stages of the patient-AI interaction lifecycle. 
In contrast, Healthbench shows a concentration in Professional Medical tasks (45.2\%), aligning with its focus on evaluating advanced medical expertise and professional knowledge.

\paragraph{Top 15 Specific Labels (Right Panel)}
The right panel reveals the most frequent specific task types across both datasets. Knowledge Query represents the dominant category in QuarkMedBench (17.8\%), followed by Symptom Analysis (12.5\%) and Treatment Guidance (11.2\%). This granular distribution reflects common intents in real-world patient queries. Conversely, Healthbench shows elevated frequencies in specialized diagnostic tasks (e.g., Differential Diagnosis at 8.7\%), reinforcing its utility in evaluating specialist reasoning capabilities.

\paragraph{Synthesis and Implications}
Taken together, the scenario-level and task-level analyses demonstrate that Healthbench and QuarkMedBench are optimized for different evaluation objectives. Healthbench is highly rigorous for evaluating advanced clinical reasoning and professional medical knowledge. QuarkMedBench, on the other hand, provides enhanced coverage of high-prevalence outpatient conditions, preventive health topics, and localized contexts such as TCM. By maintaining a balanced distribution across the medical consultation lifecycle, QuarkMedBench offers an ecologically valid benchmark for developing and evaluating AI health assistants intended for patient-facing deployment and daily health management.

\
\end{document}